\documentclass[letterpaper]{article} % DO NOT CHANGE THIS
\usepackage{aaai2026}  % DO NOT CHANGE THIS
\usepackage{times}  % DO NOT CHANGE THIS
\usepackage{helvet}  % DO NOT CHANGE THIS
\usepackage{courier}  % DO NOT CHANGE THIS
\usepackage[hyphens]{url}  % DO NOT CHANGE THIS
\usepackage{graphicx} % DO NOT CHANGE THIS
\usepackage{natbib}  % DO NOT CHANGE THIS AND DO NOT ADD ANY OPTIONS TO IT
\usepackage{caption} % DO NOT CHANGE THIS AND DO NOT ADD ANY OPTIONS TO IT
\usepackage{algorithm}
\usepackage{algorithmic}

\usepackage{newfloat}
\usepackage{listings}
\DeclareCaptionStyle{ruled}{labelfont=normalfont,labelsep=colon,strut=off} % DO NOT CHANGE THIS
\floatstyle{ruled}
\newfloat{listing}{tb}{lst}{}
\floatname{listing}{Listing}
\usepackage{multirow}
\usepackage{enumitem}
\usepackage{makecell}
\usepackage{amsmath}
\usepackage{amssymb}
\usepackage{graphicx}
\usepackage{subcaption}
\usepackage{comment}
\usepackage{diagbox}
\usepackage{booktabs}
\usepackage{cleveref}
\usepackage{arydshln}

\newcommand{\Tref}[1]{Table~\ref{#1}}

\newcommand{\Eref}[1]{Eq.~(\ref{#1})}

\newcommand{\Fref}[1]{Figure~\ref{#1}}

\newcommand{\Sref}[1]{Section~\ref{#1}}

\newcommand{\myparagraph}[1]{\vspace{6pt}\noindent\textbf{#1}\hspace{0.5em}}

\def\naive{na\"ive\ }
 
\def\ie{\emph{i.e.}}

\def\ourPaperTitle {4D Scaffold Gaussian Splatting with Dynamic-Aware Anchor Growing for Efficient and High-Fidelity Dynamic Scene Reconstruction}

\title{\ourPaperTitle}
\author{
    Woong Oh Cho,
    In Cho,
    Seoha Kim,
    Jeongmin Bae,
    Youngjung Uh,
    Seon Joo Kim
}
\affiliations{
    Yonsei University\\
    \{wocho, join, hailey07, jaymin.bae, yj.uh, seonjookim\}@yonsei.ac.kr
}

\usepackage{bibentry}
\begin{document}

\maketitle

% \begin{figure}
%     \centering
%     \vspace{-1em}
%     \includegraphics[width=0.98\linewidth]{Main/Assets/figure_teaser_v2_a.pdf}
%     \vspace{1em}
%     (a) Fiedlity-storage trade-off
%     \includegraphics[width=0.98\linewidth]{Main/Assets/figure_teaser_v2_b.pdf}
%     (b) Effectiveness of proposed method
%     \caption{
%     \textbf{Teaser.}
%     %%% Ours 라는 표헌 x
%     (a) \needrevise{Ours} achieves superior fidelity on reconstructed dynamic regions and low storage compared to other 4D Gaussian methods. Size of circle means higher FPS. 
%     (b) Our temporal-aware anchor growing and modified temporal opacity (\textbf{Ours}) improves the proposed implementation of 4D anchor-based Gaussian method (\textbf{Naive 4D Scaffold}).
%     }
%     \label{fig:teaser}
%     \vspace{-2mm}
% \end{figure}

\begin{figure*}
    \centering
    \includegraphics[width=\linewidth]{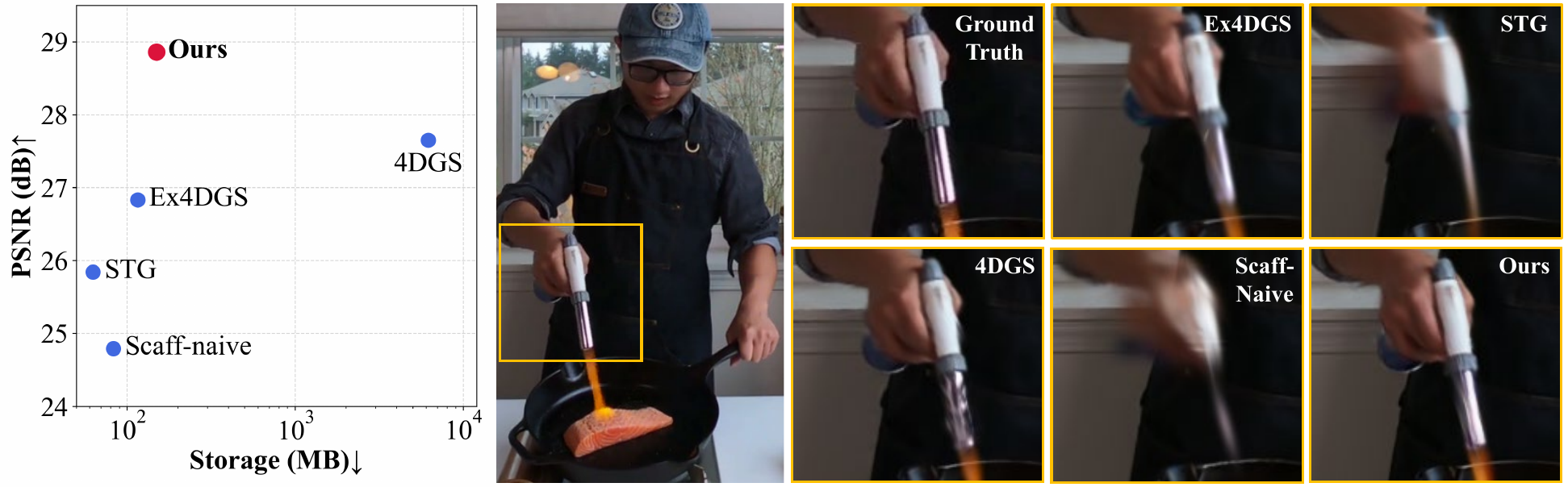}
    \caption{
    \textbf{Dynamic region quality comparison.}
    \textbf{(left)} Comparison of dynamic region quality ($y$-axis) versus storage cost ($x$-axis) on N3DV. \textbf{(right)} Results on the \textit{flame\_salmon} scene from N3DV.
    Scaff-naive refers to the naive anchor-based model without our neural Gaussian design and dynamic-aware anchor growing.
    Our method significantly outperforms all baselines with efficient storage usage, while other methods either exhibit degraded quality in dynamic regions or excessive storage costs.
    }
    \label{fig:teaser}
    \vspace{-8pt}
\end{figure*}

% \begin{center}
%     \vspace{-8mm}
%     \includegraphics[width=\linewidth]{Main/Assets/figure_teaser_v3.pdf}
%     % \vspace{-6mm}
%     % \captionsetup{skip=2pt}
%     \captionof{figure}{
%     \textbf{Teaser.} Our method achieves superior image quality and low storage surpassing other 4D Gaussian methods.}
%     \label{fig:teaser}
% \end{center}

% \begin{figure*}[htbp] % Use figure* for full-width figure in two-column mode, or figure for one-column
%     \centering
%     \begin{subfigure}[b]{0.34\textwidth} % Adjust width as needed
%         \centering
%         \includegraphics[width=\textwidth]{Main/Assets/figure_plot_teasor.pdf} % Path to your first PDF file
%         \caption{Description for (a)}
%         \label{fig:teaser_a}
%     \end{subfigure}
%     \hfill % This command pushes the subfigures apart
%     \begin{subfigure}[b]{0.625\textwidth} % Adjust width as needed, ensure total width < \textwidth
%         \centering
%         \raisebox{0.85cm}{
%         \includegraphics[width=\textwidth]{Main/Assets/figure_teaser_v4b.pdf} % Path to your second PDF file
%         }
%         \caption{Description for (b)}
%         \label{fig:teaser_b}
%     \end{subfigure}
%     \caption{Overall caption for both figures (a) and (b).}
%     \label{fig:teaser}
% \end{figure*}
\begin{abstract}
Modeling dynamic scenes through 4D Gaussians offers high visual fidelity and fast rendering speeds, but comes with significant storage overhead.
Recent approaches mitigate this cost by aggressively reducing the number of Gaussians.
However, this inevitably removes Gaussians essential for high-quality rendering, leading to severe degradation in dynamic regions.
In this paper, we introduce a novel 4D anchor-based framework that tackles the storage cost in different perspective.
Rather than reducing the number of Gaussians, our method retains a sufficient quantity to accurately model dynamic contents, while compressing them into compact, grid-aligned 4D anchor features.
%%%% [in] 이 문장은 디테일한거같기도? 뺄지말지 고민
Each anchor is processed by an MLP to spawn a set of neural 4D Gaussians, which represent a local spatiotemporal region.
We design these neural 4D Gaussians to capture temporal changes with minimal parameters, making them well-suited for the MLP-based spawning.
Moreover, we introduce a dynamic-aware anchor growing strategy to effectively assign additional anchors to under-reconstructed dynamic regions.
Our method adjusts the accumulated gradients with Gaussians' temporal coverage, significantly improving reconstruction quality in dynamic regions.
Experimental results highlight that our method achieves state-of-the-art visual quality in dynamic regions, outperforming all baselines by a large margin with practical storage costs.
\end{abstract}

% Uncomment the following to link to your code, datasets, an extended version or similar.
% You must keep this block between (not within) the abstract and the main body of the paper.
% \begin{links}
%     \link{Code}{https://aaai.org/example/code}
%     \link{Datasets}{https://aaai.org/example/datasets}
%     \link{Extended version}{https://aaai.org/example/extended-version}
% \end{links}

\section{Introduction}

Reconstructing dynamic scenes from multi-view videos has received significant attention due to its wide applications.
Beyond methods based on neural radiance fields (NeRFs) \cite{li2022neural, wang2022mixed, attal2023hyperreel}, 
3D Gaussian Splatting (3DGS) \cite{kerbl3Dgaussians} has become a major approach for dynamic scene reconstruction, due to its ability to render high-quality novel views in real-time.
The two main categories of this approach are 1) modeling temporal changes as deformations of canonical 3D Gaussians \cite{yang2023deformable3dgs, bae2024pergaussian, xu2024grid4d} and 2) employing 4D Gaussians to approximate a scene's 4D volumes \cite{yang2023gs4d, li2023spacetimegaussians, Lee_2024_C3DGS, lee2024ex4dgs}.

Directly optimizing 4D Gaussians offers higher visual quality and faster rendering than deformation-based methods.
By modeling temporal changes through multiple 4D Gaussians, each covering a certain time range, they effectively capture complex dynamic regions without the heavy computation cost of deformation fields.
However, this approach accompanies a large number of 4D Gaussians, which subsequently leads to substantial storage overhead--often exceeding 6GB for a 10-second video (see \Fref{fig:teaser}, left).

%%% [+] 4DGS storage를 해결하기 위해 efficient 4D methods 등장
%%%%% -> 주로 Gaussian 갯수를 줄여서 용량 확보하려 함
%%%%% -> 스토리지 성능은 잘 줄었고, 전체 성능은 잘 나오는 것 보이지만, complex dynamic region에서는 장렬히 실패
%%%%% -> 이들의 스토리지 성능은, rendering quality 특히 dynamic regions의 성능을 희생해서 달성된 것
Several approaches have attempted to address the storage cost by reducing the number of Gaussians.
They achieve this by more complicated motion modeling \cite{lee2024c3dgs}, aggressive pruning \cite{li2023spacetimegaussians}, or motion interpolation \cite{lee2024ex4dgs}.
While these methods effectively reduce storage, they also inevitably remove Gaussians in dynamic regions, sacrificing the expressiveness needed to represent complex temporal changes.
As a result, these methods often suffer from quality degradation in dynamic components.

In this paper, we introduce a 4D anchor-based framework that addresses the storage overhead from a different perspective.
Instead of reducing the number of Gaussians--which is crucial for maintaining the expressiveness--our method maintains a sufficient number to render dynamic regions with high visual quality.
This is achieved by representing dynamic scenes through structured anchor features \cite{lu2024scaffold}, aligned with a sparse 4D grid.
Each anchor holds a compact feature vector to model a local spatiotemporal region.
This feature is processed by shared multi-layer perceptrons (MLP) to output properties of neural Gaussians.

%%% [+] Our scaffold scheme은 anchor feature -> MLP -> Gaussian 생성
%%%%% MLP output이기 때문에, Gaussian property를 simple yet effective하게 디자인해야함
%%%%% Gaussian의 dynamic을 두 가지 component(?)로 표현
%%%%% 1. simplifies slicing of 4DGS and models movements through liner motions
%%%%% 2. models temporal opacity through generalized Gaussian -> sudden changes 더 잘 모델링
% 웅오 하이! 웅오 화이팅~~~~~~~
%%%%% 이 두개를 합쳐서, 우리는 dynamic을 set of piecewise linear segments로 표현
Considering the limited capacity of the shared shallow MLP, we propose a compact parametrization of the neural Gaussians that effectively captures temporal changes.
This includes two key modelings: linear motion to model the time-varying 3D position, and a generalized Gaussian function to model temporal opacity, which is better suited for capturing sudden appearance changes.
This design enables our framework to effectively represent complex dynamic trajectories through a sequence of linear segments.

%%% limitation of naive extension
While the anchor-based scheme effectively reduces storage, a \naive extension of 3D scaffolding \cite{lu2024scaffold} fails to accurately capture dynamic regions (see \Fref{fig:teaser}, Scaff-naive).
The limitation arises from the anchor growing strategy used in \cite{lu2024scaffold}, which is originally designed for static scenes.
In our 4D framework, each anchor and its neural Gaussians are active only within a specific time range, while the anchor growing strategy accumulates gradients across all frames without considering this coverage.
As a result, dynamic regions appearing in only a few frames receive less gradient signals, leading to under-reconstructed dynamic components in the final scene.

%%% dynamic anchor growing
We present a dynamic-aware anchor growing strategy to properly allocate new anchors to under-reconstructed dynamic regions.
Our approach adjusts the accumulated gradients based on each Gaussian's temporal coverage.
We adjust dynamic Gaussians with shorter coverage to have larger gradient, compensating for the penalty caused by their short appearance.
This modification encourages new anchors to be allocated more in under-reconstructed dynamic regions, achieving substantial improvement in reconstruction quality.

Our anchor-based framework, coupled with the dynamic-aware anchor growing strategy, represents complex dynamic components in high-quality while addressing storage costs.
We validate our approach through extensive experiments conducted on the N3DV \cite{li2022neural} and Technicolor \cite{sabater2017dataset} datasets.
Notably, our method outperforms all baselines by a large margin with practical storage overhead, as illustrated in \Fref{fig:teaser}.
% [웅] 윗 두줄은 같은 얘기를 반복하고 있는것 같은데, 한 줄로 줄여도 되지 않을까?
\section{Related Work}
% This section presents a comprehensive review of scene representations, focusing on three aspects. First, we examine approaches for efficient 3D Gaussians (\cref{relwork:efficient_3dgs}). Then, we investigate methods for both deforming 3D Gaussians and utilizing 4D Gaussians in dynamic scene reconstruction (\cref{relwork:dynamic_3dgs}). Lastly, we review various approaches employing neural features for scene reconstruction (\cref{relwork:feature}). 
% the methods for feature-based compression that enable efficient scene reconstruction.

\begin{figure*}
    \centering
    \includegraphics[width=1\linewidth]{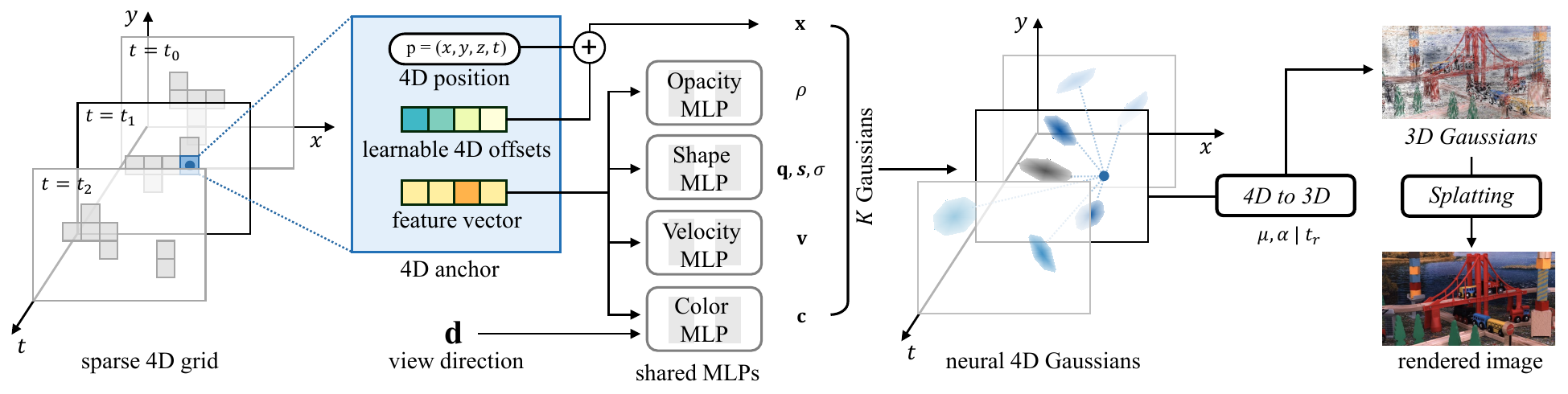}
    \caption{
    \textbf{Proposed 4D anchor-based framework.} We begin with a sparse 4D grid of anchors, each defined by a unique 4D spatiotemporal position $\mathbf{p}$ and learnable offsets. Shared MLPs utilize these anchors to generate neural 4D Gaussians, capturing dynamic appearance changes with view direction $\mathbf{d}$. These 4D Gaussians are then projected to 3D Gaussians at specific time $t_r$ for rendering, using a 3D Gaussian splatting pipeline to produce the final rendered image. We omit the $z$-axis for simplicity.
    }
    % \vspace{-2mm}
    \label{fig:method}
\end{figure*}
\myparagraph{Efficient 3D Gaussians.}
In recent years, 3D Gaussian Splatting (3DGS) has attracted significant attention for achieving real-time rendering by representing scenes with 3D Gaussian primitives and introducing a tile-based rasterizer. 
For photorealistic rendering quality, 3DGS requires a significant number of 3D Gaussians, which increases storage costs. 
To address this problem, previous works remove unnecessary Gaussians that do not harm rendering quality \cite{fan2023lightgaussian, girish2023eagles}, replace Gaussians or their parameters with efficient representations \cite{niedermayr2024compressed, papantonakis2024reducing}, and compress Gaussian parameters using existing compression techniques such as entropy coding and image compression \cite{niedermayr2024compressed, chen2025hac, xie2025mesongs}.
In this paper, we address compressing Gaussians in spatiotemporal space, which have been relatively unexplored compared to 3D Gaussians.

\vspace{-2pt}

\myparagraph{Dynamic 3D Gaussians.}
Two main approaches are proposed to extend 3DGS \cite{kerbl3Dgaussians} into dynamic scene reconstruction. The first approach involves deforming 3D Gaussians along with temporal changes \cite{yang2023deformable3dgs, wu20234dgaussians, bae2024pergaussian, shaw2024swings, xu2024grid4d, kwak2025modecgs}. 
These deformable 3D Gaussians offer the advantage of compact storage requirements but exhibit relatively slow rendering speeds and low visual quality.
In contrast, the other approach directly employs 4D Gaussians in the spatio-temporal domain.
4DGS \cite{yang2023gs4d} demonstrate superior visual quality and faster rendering speeds, but suffer from higher storage requirements.
Although some works \cite{li2023spacetimegaussians, lee2024ex4dgs, Lee_2024_C3DGS} improve storage efficiency using fewer Gaussians, they tend to focus on the holistic scene, thereby neglecting the quality of dynamic areas.
To address this, we take an alternative approach: reducing storage while maintaining quality by preserving the Gaussian count.

\myparagraph{Feature-based neural rendering.}
\label{relwork:feature}
Recently, a growing trend in scene reconstruction has been to integrate neural features as additional inputs to enhance model performance. For instance, there have been attempts to extract features from source views and utilize them for novel view synthesis, enabling few-view reconstruction \cite{yu2021pixelnerf, mvsnerf} or view interpolation through transformers \cite{wang2021ibrnet, reizenstein2021commonobjects3dlargescale, t2023is}. 
In 3DGS studies, 
Compact3DGS \cite{deng2024compact} uses a hash grid instead of per-Gaussian SH coefficients and
STG \cite{li2023spacetimegaussians} renders features into RGB via shallow MLPs. Some works generate Gaussian attributes from a multi-level tri-plane \cite{wu2024implicit} and predict the attributes of local 3D Gaussians from anchor features \cite{lu2024scaffold}. We introduce a 4D anchor-based framework that includes dynamic linear motion and temporal opacity derived from a generalized Gaussian distribution, considering real-world dynamics.
\section{Method}
% Reconstructing dynamic 3D scenes with 4D Gaussians offer more flexibility for capturing dynamics, especially in rapidly changing regions.
% This results in better visual quality and faster rendering speeds without necessitating expensive computations of per-Gaussian deformations.

We reconstruct dynamic 3D scenes through a sparse, grid-aligned 4D anchor grid.
Each anchor holds a compressed feature vector that specifies nearby 4D Gaussians.
The dynamic-aware anchor growing strategy encourages new anchors to be closely placed at under-reconstructed dynamic regions.
Furthermore, we design each 4D Gaussian to move along a line segment, and define the temporal opacity function which fits a piecewise persistent period.
Through these components, our framework employs a sufficient number of Gaussians to achieve high-quality reconstruction of dynamic regions, while the storage overhead is significantly reduced via compressed anchor features.

\begin{comment}
    
We introduce a 4D anchor-based framework for dynamic 3D scene reconstruction, which effectively and efficiently models dynamic appearance changes in scenes.
The main goal of our method is resolving redundant patterns present across space and time through grid-aligned 4D anchors.
% , while exploiting flexibility of 4D Gaussians
We first illustrate the overview of our method in \Sref{sec:method_overview}. Then we present the temporal coverage-aware anchor growing strategy in \Sref{sec:method_growing}, and the formulations of neural 4D Gaussians, including the neural velocity and the temporal opacity in \Sref{sec:method_4d}.
\end{comment}

\subsection{4D Anchor-Based Framework}
\label{sec:method_overview} 
The overview of our method is illustrated in \Fref{fig:method}.
Our method begins with a set of sparse 4D anchor points, each of which has a unique, grid-aligned spatiotemporal 4D position $\mathbf{p} \in \mathbb{R}^4$ with a feature vector $\mathbf{f} \in \mathbb{R}^C$.
We leverage shared MLPs to produce $K$ neural 4D Gaussians from these anchor features.
Corresponding 3D Gaussians are computed from these 4D Gaussians to render a frame at timestep $t$.
%We describe details of our method in the following paragraphs.

% \textblock{Initializing 4D anchors.}
\myparagraph{Initializing 4D anchors.}
Similar to 3D Scaffold-GS \cite{scaffoldgs}, we utilize static point clouds to initialize 4D anchor points.
We first obtain the static point cloud from multi-view frames at a certain timestep $t_{0}$ using Structure-from-Motion (SfM) \cite{schoenberger2016sfm}.
%%% [in] 뒤에 velocity에서 mathbf{v} 사용 .. 
The 4D positions of anchors are initialized from a set of voxels $\mathbf{V} \in \mathbb{R}^{N\times3}$ obtained from this point cloud:
\begin{equation}
    \mathbf{p} = (x_v, y_v, z_v, t_{0}),\ \ \forall v \in \mathbf{V},
\end{equation}
\begin{equation}
    \mathbf{x}_k = \mathbf{p} + \Delta\mathbf{x}_k,\ \ k \in \{1, 2, ..., K\},
\end{equation}
where ($x_v$, $y_v$, $z_v$) is the center coordinates of the voxel $v$.
Each anchor point also accompanies two learnable parameters: a feature vector $\mathbf{f}$, and 4D offsets $\Delta\mathbf{x}_k \in \mathbb{R}^{4}$ for determining properties of $K$ neural 4D Gaussians. 4D Gaussian position $\mathbf{x}$ is determined by anchor position and offset.
%%%% [in] parameter init -> implementation detail로 이동 예정
%%%%%% [w] position -> properties

%%% [in] 넣을지 말지...
% Beginning with 4D anchors initialized with the static point cloud, our method gradually increases anchors to dynamic regions.
% This is accomplished by our anchor growing strategy, which will be described in dynamic-aware anchor growing.

% \textblock{Neural 4D Gaussians.}
\myparagraph{Neural 4D Gaussians.}
We leverage shared MLPs and learnable parameters of the 4D anchors to generate $K$ neural 4D Gaussians from each anchor.
Specifically, shared MLPs take the anchor feature $\mathbf{f}$ and yield properties of $K$ neural Gaussians, which include base opacity $\rho \in \mathbb{R}$, quaternion $\mathbf{q} \in \mathbb{R}^4$ and scaling $\mathbf{s} \in \mathbb{R}^3$ for the covariance matrix, view-dependent color $\mathbf{c} \in \mathbb{R}^3$.
Our shared MLPs also produce a temporal scale $\sigma \in \mathbb{R}$ to compute our temporal opacity, and the neural velocity $\mathbf{u} \in \mathbb{R}^3$.
Note that the color MLP additionally takes the view direction $\mathbf{d} \in \mathbb{R}^3$ as inputs for view-dependent modeling.

%%% [in] or.. title: 4D to 3D Gaussians
% \textblock{Rendering neural Gaussians.}
\myparagraph{Rendering neural Gaussians.}
To render the neural 4D Gaussians, we compute 3D Gaussian parameters at time $t_r$ from our time-varying properties.
For the $k$-th Gaussian of the anchor $\mathbf{p}$, the center $\mu_k \in \mathbb{R}^3$ and the opacity $\alpha_k \in \mathbb{R}$ of the corresponding 3D Gaussian at time $t_r$ are derived as 
\begin{equation}
\label{eq:mu_3d}
    \mu_k = \mathbf{x}^{xyz}_{k} + h(t_r, \mathbf{x}^{t}_{k}, \mathbf{u}),
\end{equation}
\begin{equation}
\label{eq:opacity_3d}
    \alpha_k = \rho_k \cdot g(t_r, \mathbf{x}^{t}_{k}, \sigma_k),
\end{equation}
%%%% [in] h, g 에 대한 설명 다시 쓰기
where $\mathbf{x}^{xyz}_{k} \in \mathbb{R}^3$ and $\mathbf{x}^{t}_{k} \in \mathbb{R}$ are the spatial and the temporal position of the 4D Gaussian. 
$h(\cdot)$, $g(\cdot)$ model time-varying values of the positions and the opacity with the neural velocity and temporal opacity function, which will be further described in the following section.

After deriving 3D Gaussians at time $t_r$, we utilize the existing 3D Gaussian splatting pipeline \cite{kerbl3Dgaussians} to render the frames. Same as 3D Scaffold-GS \cite{scaffoldgs}, only Gaussians within the view frustum and having opacity higher than the threshold are passed to the renderer.

\subsection{Compact Parametrization of 4D Gaussians}
% [웅] Minimal Parametrization은 어떤지?
\label{sec:method_4d}
The properties of neural 4D Gaussians are predicted by a shallow shared MLP.
Increasing the number of Gaussian properties complicates the optimization process.
To this end, we propose a compact parametrization of 4D Gaussians that captures temporal changes as a set of linear segments.
Our neural Gaussian design involves two key modelings. First, we represent time-varying spatial positions as linear motions. Second, we employ a generalized Gaussian function to model temporal opacity, which effectively captures sudden appearance changes with a single parameter.

\myparagraph{Linear motion.}
% In computer graphics, one common way to model the motion of an object is through a set of piecewise linear segments.
% Based on this concept, we propose the neural velocity to model the motions of scene elements with a set of 4D Gaussians having linear movements.
% We introduce a neural velocity term that simplifies the temporal slicing of Gaussian positions, as used in 4DGS \cite{lee2024ex4dgs}.
As described in \Eref{eq:mu_3d}, the time-varying 3D position of the 3D Gaussian $\mu_k \in \mathbb{R}^3$ is determined by the time-varying function $h(\cdot)$.
We formulate the $h(\cdot)$ as a linear motion based on the neural velocity $\mathbf{u}$:
\begin{equation}
    h(t, \mathbf{x}^{t}_{k}, \mathbf{u}) = (t - \mathbf{x}^{t}_{k})\mathbf{u}.
\end{equation}
This formulation simplifies the temporal slicing of Gaussian positions proposed in 4DGS \cite{lee2024ex4dgs}, using only 3 parameters per Gaussian.
%%% [in] 분량 보고 빼든가 말든가 하자
Despite its simplicity, a set of Gaussians with linear motion effectively captures the temporal dynamics of scene elements, particularly when combined with our modified temporal opacity.

\begin{figure}
    \centering
    \includegraphics[width=0.95\linewidth]{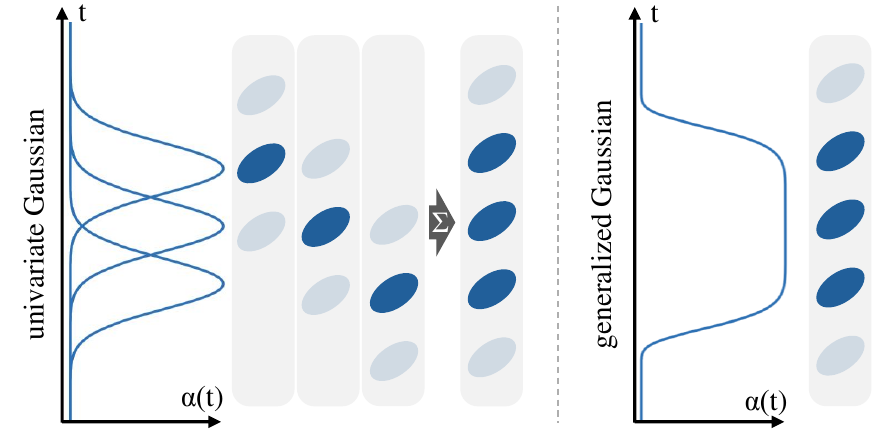}
    \caption{
    \textbf{Illustration of the modified temporal opacity.}
    To reconstruct the sudden and continuous appearance of an object, the univariate Gaussian function requires the sum of multiple Gaussians, while the generalized Gaussian achieves the same expressiveness with only a single Gaussian.
    \label{fig:method_temporal_opacity}
    }
    % \vspace{-2mm}
\end{figure}
\myparagraph{Modified temporal opacity.}
Previous 4D Gaussian methods \cite{yang2023gs4d, li2023spacetimegaussians} typically adopt the univariate Gaussian function to make Gaussians appear within a specific time range.
However, this formulation lacks the capacity to model abrupt changes in real-world scene elements, limiting the expressiveness of individual Gaussians.

We reformulate the temporal opacity function based on a generalized Gaussian distribution. This offers steeper derivatives at the beginning and the end, making it better suited to fit piecewise persistent periods.
Our time-varying opacity $g(\cdot)$ described in \Eref{eq:opacity_3d} is formulated as
\begin{equation}
    g(t, \mathbf{x}^{t}_{k}, \sigma_k) = exp({-(|t - \mathbf{x}^{t}_{k}| / \sigma_k)^\beta}),
    \label{eq:temporal-opacity}
\end{equation}
where $\beta$ is a tunable hyperparameter.
In practice, we set the inverse sigma as a learnable parameter, as it leads to more stable training.
We also model $\beta = 2\beta'$ and $\beta'$ as a hyperparameter, which removes the $|\cdot|$ in the above equation.

While the issue of temporal opacity is also explored in \cite{lee2024ex4dgs}, we show that efficiently representing temporal coverage with fewer parameters is crucial for the performance of the anchor-based scheme.

\subsection{Dynamic-Aware Anchor Growing}
\label{sec:method_growing}
Growing new anchors to under-reconstructed dynamic regions is a critical factor for high-quality results.
The direct application of previous anchor growing to the dynamic reconstruction fails to identify such regions, as it neglects the temporal coverage of the Gaussians.

The previous anchor growing strategy \cite{scaffoldgs} designed for static scenes gathers the gradients as a mean over $N$ iterations:
\begin{equation}
        \nabla_{g} = \frac{\sum^{N} \|\nabla_{\text{2D}}\|}{N}.
\end{equation}
With this strategy, dynamic regions appearing in a short period will have lower $\nabla_{g}$, as these regions will be penalized by the denominator $N$ regardless of their actual errors.

%%% [in] NEEEEED TO CHANGE NAME to dynamic-ware anchor growing
To address this, we propose a dynamic-aware anchor growing operation.
The core of our anchor growing is the computation of the accumulated gradients of each 4D Gaussian $\nabla_{g}$.
We formulate $\nabla_{g}$ as a weighted sum of the gradient over $N$ iterations:
%%%%%%%% [in] 2D gradient가 맞나..? -> [woong] 정확히는 2D position gradient
\begin{equation}
    \nabla_{g} = \frac{\sum^{N} w(\alpha', \sigma) \|\nabla_{\text{2D}}\|}{\sum^{N} w(\alpha', \sigma)},
\end{equation}
% \alpha(t) (1/\sigma)^\lambda
% with the temporal opacity $\alpha(t)$ and inverse of temporal scale $\sigma$ 
where $\nabla_{\text{2D}}$ is the 2D position gradient of the Gaussian, and $\alpha' = g(t_r, \mathbf{x}^t, \sigma)$ is the time-variant component of the Gaussian opacity.
We define the weight term $w$ as a function of $\alpha$ and $\sigma$, with $\gamma$ as a hyperparameter:
\begin{equation}
\label{eq:temporalweight}
    w(\alpha', \sigma) = \alpha'  (1/\sigma)^\gamma.
\end{equation}
We collect the accumulated gradients of all Gaussians and then voxelize them.
New anchors are placed at the centers of the voxels having $\nabla_{\text{2D}}$ higher than each threshold.

The dynamic-aware anchor growing strategy differs from the original anchor growing operation \cite{lu2024scaffold} in two key aspects.
First, gradients are accumulated only when the Gaussian's temporal opacity is greater than zero (\ie, when it is activated).
It allows our method to accurately gather the gradients of 4D Gaussians placed at the regions where dynamic scene elements only appear in a short period.
Second, the temporal coverage $\sigma$ directly influences to the anchor growing operation.
This encourages regions with a short temporal coverage to receive stronger gradients, making it easier for anchors to grow in those areas.
By accumulating gradients as a weighted sum, our method effectively places new anchors to under-reconstructed dynamic regions.

\subsection{Training and Implementation Details}

\begin{table*}[t]
    \setlength{\tabcolsep}{2pt} % Adjust column separation
    \centering
    % \resizebox{1.7\columnwidth}{!}{
    \resizebox{1.0\linewidth}{!}{ 
    \begin{tabular}{l|cccc|cccc|ccc}
        \toprule
        & \multicolumn{4}{c|}{Dynamic region} & \multicolumn{4}{c|}{Full region} & \multicolumn{3}{c}{Computational cost} \\
        \textbf{Model} & \textbf{PSNR$\uparrow$} & \textbf{SSIM$\uparrow$} & \textbf{LPIPS$\downarrow$} & \textbf{\#Gaussians} & \textbf{PSNR$\uparrow$} & \textbf{SSIM$\uparrow$} & \textbf{LPIPS$\downarrow$} & \textbf{\#Gaussians} & \textbf{FPS$\uparrow$} & \textbf{Storage$\downarrow$} & \textbf{Training time$\downarrow$}  \\
        \toprule
        % \midrule
        % for the storage and fps, we use reported value in ed3dgs, they only include the salmon1 scene 
        % K-Planes \cite{kplanes_2023} & 31.22 & \underline{0.947} & 0.090 & a & b & c & 1 h 40 m & 309 MB & 0.13 \\ % K-Planes Hybrid
        % MixVoxels-L \cite{wang2022mixed} & 30.81 & 0.933 & 0.095 & a & b & c & 1 h 40 m & 512 MB & 0.93 \\ % MixVoxels L
        % HyperReel \cite{attal2023hyperreel} $^1$ & 31.10 & 0.927 & 0.099 & a & b & c & 9 h 20 m & 1362 MB & 1.04 \\ % split 300 frames into 50 frames and train each model
        % \hline
        4DGaussian \shortcite{wu20234dgaussians} & 25.33 & 0.833 & 0.166 & - & 30.71 & 0.935 & 0.056 & 192 K & 51.9 & 57 MB & 50 m \\
        E-D3DGS \shortcite{bae2024pergaussian} & 26.92 & 0.884 & 0.112 & - & \underline{32.04} & \textbf{0.951} & \textbf{0.034} & 180 K & 74.5 & 66 MB & 1 h 52 m  \\
        Grid4D \shortcite{xu2024grid4d}  & 26.65 & 0.877 & 0.129 & - & 31.92 & \underline{0.949} & \underline{0.039} & 202 K & 127.1 & 48 MB & 1 h 20 m \\
        \midrule
        STG \shortcite{li2023spacetimegaussians} & 25.84 & 0.860 & 0.127 & 44 K & 31.24 & 0.941 & 0.051 & 434 K & 125.9 & 63 MB & 1 h 8 m \\
        4DGS \shortcite{yang2023gs4d} &  \underline{27.65} & \underline{0.907} & \underline{0.075} & 3306 K & \textbf{32.14} & 0.947 & 0.047 & 3333 K & 61.4 & 6194 MB & 9 h 30 m  \\
        4DGS$^\dagger$ (1GB) & 26.70 & 0.877 & 0.123 & 557 K & 31.59 & 0.943 & 0.052 & 581 K & 152.7 & 1080 MB & 3 h 37 m  \\
        % STG \cite{li2023spacetimegaussians} $^1$ & \underline{29.26}             & \textbf{0.934} & \underline{0.051} & 31.96 & \underline{0.948} & 0.046 & 125.9 & - MB & - K & 5 hours \\
        % .C3DGS \shortcite{Lee_2024_C3DGS} & - & - & - & - K & - & - & - & - & - & - & - \\
        Ex4DGS \shortcite{lee2024ex4dgs} & 26.33 & 0.874 & 0.121 & 52 K & 32.01 & 0.947 & 0.048 & 268 K & 91.9 & 115 MB & 1 h 6 m \\
        \midrule
        Scaff-naive & 24.79 & 0.811 & 0.199 & 206 K & 31.21 & 0.942 & 0.053 & 732 K & 159.7 & 83 MB & 2 h 57 m \\
        Ours-light & 27.50 & 0.902 & 0.076 & 181 K & 31.54 & 0.944 & 0.045 & 314 K & 148.2 & 90 MB & 2 h 51 m \\
        Ours & \textbf{28.86} & \textbf{0.927} & \textbf{0.054} & 533 K & 32.03 & 0.947 & 0.041 & 775 K & 129.9 & 149 MB & 3 h 6 m \\
        \bottomrule
    \end{tabular}
    }
    \caption{
    \textbf{Quantitative results on the N3DV dataset.} 
    % $^1$ only includes the \textit{flame\_salmon} for storage and FPS \cite{bae2024pergaussian}. 
    % $^1$ STG divides the scene into 50 frames and trains each using a separate model and uses COLMAP \cite{schoenberger2016sfm} for all 300 frames.
    4DGS$^\dagger$ refers to the result with 1GB storage, for fair comparisons.
    % \memo{add description}
    }
    % \vspace{-2mm}
    \label{table:comparison_n3dv}
\end{table*}

\begin{figure*}[ht]
    \centering
    \includegraphics[width=0.98\linewidth]{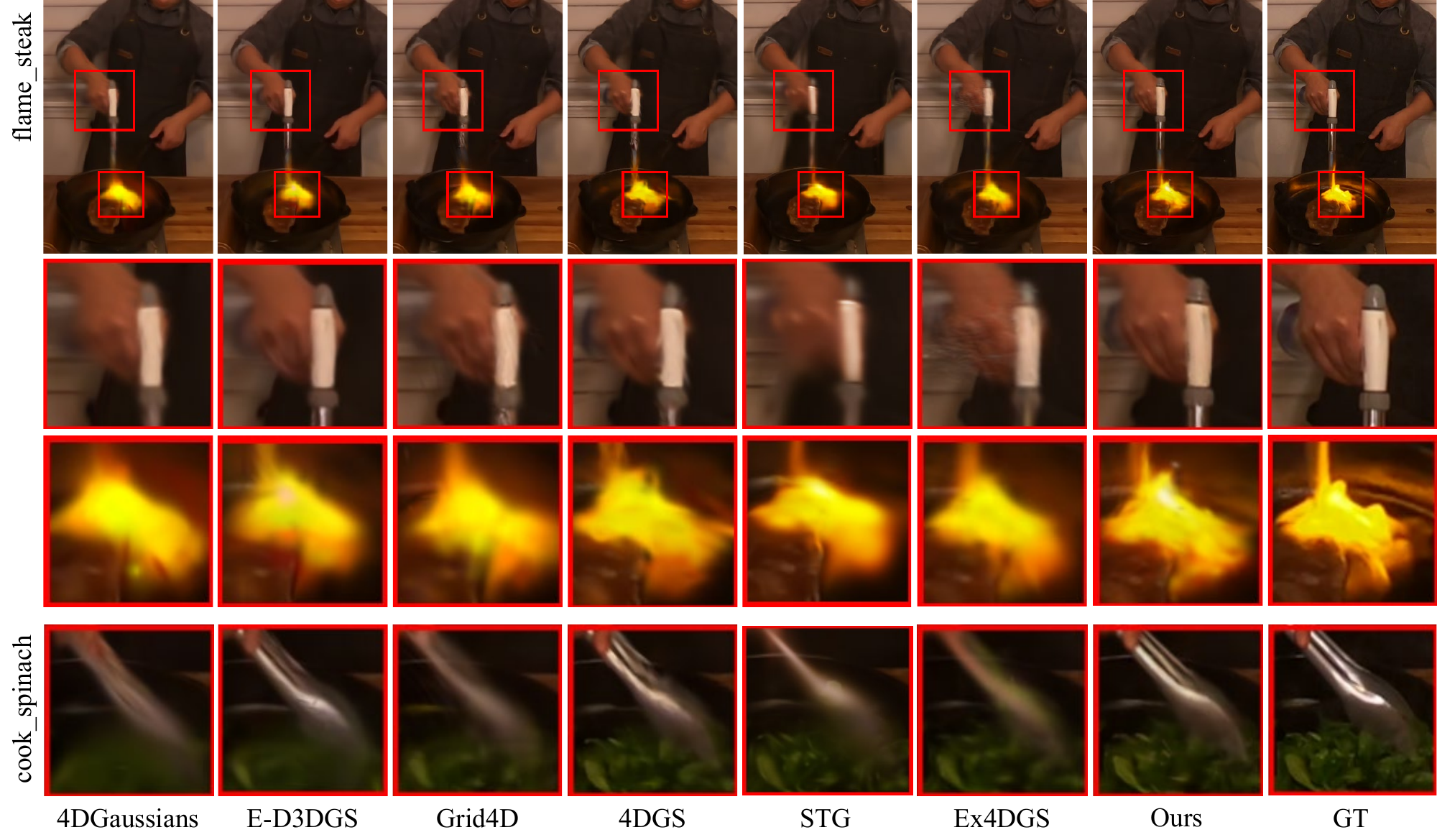}
    \caption{\textbf{Qualitative comparisons on the N3DV dataset.} Our method achieves high-fidelity in both static and dynamic regions.}
    % \vspace{-2mm}
    \label{fig:comparison_n3dv}
\end{figure*}

% \textblock{Training objective.}
\myparagraph{Training objective.}
%%%% [in] 내용 확인 -> [woong] 수식 수정 함
Both learnable anchor parameters and MLP weights are jointly optimized through the rendering loss.
We employ $\mathcal{L}_{1}$ with SSIM loss $\mathcal{L}_\text{SSIM}$ and volume regularization $\mathcal{L}_\text{vol}$ following the 3D Scaffold-GS \cite{lu2024scaffold}. 
The full training objective with weighting coefficients $\lambda_\text{SSIM}=0.2$ and $\lambda_\text{vol}=0.01$ is
% (concision) The training objective of our pipeline can be expressed as
\begin{equation}
    \mathcal{L} = (1 - \lambda_\text{SSIM})\mathcal{L}_1 + \lambda_\text{SSIM}\mathcal{L}_\text{SSIM} + \lambda_\text{vol}\mathcal{L}_\text{vol}.
\end{equation}

% \textblock{Implementation details.}
\myparagraph{Implementation details.}
Similar to 4DGS, our model is trained for 120K iterations with a single batch, which takes approximately 3 hours on a single NVIDIA A6000 GPU. We set $\beta=2$ and $\gamma=1$ for our modified temporal opacity model and dynamic-aware densification. For 4D voxel grid size, we use 0.001 as the spatial grid size and use the frame interval time as the temporal grid size for all scenes.
We set $K=10$ following \cite{lu2024scaffold}.
After training, we prune the invalid anchor points of which all Gaussians have negative opacity.
To improve inference speed, we cache the time-invariant and view-invariant outputs of MLPs.
We provide more implementation details in the supplements.

\begin{table*}[t]
    \setlength{\tabcolsep}{2pt} % Adjust column separation
    \centering
    % \resizebox{1.7\columnwidth}{!}{
    \resizebox{1.0\linewidth}{!}{ 
    \begin{tabular}{l|cccc|cccc|ccc}
        \toprule
        & \multicolumn{4}{c|}{Dynamic region} & \multicolumn{4}{c|}{Full region} & \multicolumn{3}{c}{Computational cost} \\
        \textbf{Model} & \textbf{PSNR$\uparrow$} & \textbf{SSIM$\uparrow$} & \textbf{LPIPS$\downarrow$} & \textbf{\#Gaussians} & \textbf{PSNR$\uparrow$} & \textbf{SSIM$\uparrow$} & \textbf{LPIPS$\downarrow$} & \textbf{\#Gaussians} & \textbf{FPS$\uparrow$} & \textbf{Storage$\downarrow$} & \textbf{Training time$\downarrow$}  \\
        \toprule
        % \midrule
        % for the storage and fps, we use reported value in ed3dgs, they only include the salmon1 scene 
        % K-Planes \cite{kplanes_2023} & 31.22 & \underline{0.947} & 0.090 & a & b & c & 1 h 40 m & 309 MB & 0.13 \\ % K-Planes Hybrid
        % MixVoxels-L \cite{wang2022mixed} & 30.81 & 0.933 & 0.095 & a & b & c & 1 h 40 m & 512 MB & 0.93 \\ % MixVoxels L
        % HyperReel \cite{attal2023hyperreel} $^1$ & 31.10 & 0.927 & 0.099 & a & b & c & 9 h 20 m & 1362 MB & 1.04 \\ % split 300 frames into 50 frames and train each model
        % \hline
        4DGaussian \shortcite{wu20234dgaussians} & 23.99 & 0.709 & 0.280 & - & 29.62 & 0.843 & 0.176 & 268 K & 23.4 & 72 MB & 31 m \\
        E-D3DGS \shortcite{bae2024pergaussian} & 29.39 & 0.886 & 0.148 & - & 33.38 & 0.907 & 0.100 & 212 K & 60.8 & 77 MB & 2 h 34 m  \\
        % Grid4D \shortcite{xu2024grid4d}  & 26.65 & 0.877 & 0.129 & - & 31.92 & \textbf{0.949} & \underline{0.039} & 202 K & 127.1 & 48 MB & 1 h 20 m \\
        \midrule
        STG \shortcite{li2023spacetimegaussians} & 28.35 & 0.869 & 0.174 & 71 K & 33.33 & 0.912 & 0.096 & 132 K & 149.0 & 30 MB & 1 h 24 m \\
        4DGS \shortcite{yang2023gs4d} & \underline{31.91} & \underline{0.930} & \textbf{0.093} & 5597 K & 33.30 & 0.910 & 0.095 & 5759 K & 141.5 & 10699 MB & 6 h 11 m  \\
        4DGS$^\dagger$ (1GB) & 29.18 & 0.890 & 0.144 & 538 K & 28.72 & 0.856 & 0.166 & 591 K & 156.6 & 1098 MB & 3 h 36 m  \\
        % STG \cite{li2023spacetimegaussians} $^1$ & \underline{29.26}             & \textbf{0.934} & \underline{0.051} & 31.96 & \underline{0.948} & 0.046 & 125.9 & - MB & - K & 5 hours \\
        C3DGS \shortcite{Lee_2024_C3DGS} & 27.21 & 0.843 & 0.209 & 87 K & 32.52 & 0.901 & 0.110 & 122 K & 178.7 & 18 MB & 1 h 9 m \\
        Ex4DGS \shortcite{lee2024ex4dgs} & 30.58 & 0.918 & \underline{0.109} & 120 K & 33.40 & 0.914 & 0.088 & 426 K & 78.4 & 140 MB & 1 h 53 m \\
        \midrule
        Scaff-naive & 28.00 & 0.853 & 0.195 & 721 K & 33.34 & 0.915 & 0.081 & 2056 K & 127.0 & 129 MB & 1 h 59 m\\
        Ours-light & 30.95 & 0.916 & 0.117 & 186 K & \underline{33.97} & \underline{0.923} & \
        \underline{0.074} & 480 K & 131.4 & 108 MB & 1 h 40 m \\
        Ours & \textbf{31.94} & \textbf{0.932} & \textbf{0.093} & 1165 K & \textbf{34.11} & \textbf{0.925} & \textbf{0.072} & 1272 K & 110.8 & 278 MB & 2 h 30 m \\
        \bottomrule
    \end{tabular}
    }
    \caption{
    \textbf{Quantitative results on Technicolor dataset.} 
    % $^1$ refers to the values reported in \cite{li2023spacetimegaussians}. 
    % $^1$ only includes the \textit{Painter} for storage and FPS \cite{bae2024pergaussian}. 
    % $^1$ STG uses COLMAP \cite{schoenberger2016sfm} for all 50 frames.
    For fair comparison, 4DGS$^\dagger$ result of using 1GB storage is reported.
    }
    % \vspace{-2mm}
    \label{table:comparison_technicolor}
\end{table*}

\section{Experiments}
In this section, we first evaluate our method through the comparisons with several state-of-the-art baselines for dynamic scene reconstruction.
Then we conduct ablations and analysis to explore the effectiveness of our main features.
Our code will be made publicly available.

% \subsection{Experimental setup}
% \textblock{Datasets.}
\myparagraph{Datasets.}
We evaluate our method on two representative real-world datasets, which are Neural 3D Video (N3DV)~\cite{li2022neural} and Technicolor~\cite{sabater2017dataset}.
\begin{itemize}
    \item \textbf{Neural 3D Video dataset (N3DV)} comprises 17 to 21 synchronized videos of six scenes, with each video containing 300 frames. Following previous works, we downsample the resolution to $1352 \times 1014$ and use the first camera as the test view. We exclude \textit{cam13} for the \textit{coffee\_martini} scene due to synchronization issues.
    \item \textbf{Technicolor dataset} contains 16 synchronized videos across five scenes, with each video comprising 50 frames. We retain the original resolution of $2048 \times 1088$ and use \textit{cam10} as the test view.
\end{itemize}

% \textblock{Baselines.}
\myparagraph{Baselines.}
We choose the following state-of-the-art competitors: 4DGaussian \cite{wu20234dgaussians}, E-D3DGS \cite{bae2024pergaussian} and Grid4D \cite{xu2024grid4d} representing deformation-based methods, and 4DGS \cite{yang2023gs4d}, STG \cite{li2023spacetimegaussians}, C3DGS \cite{Lee_2024_C3DGS} and Ex4DGS \cite{lee2024ex4dgs} representing 4D Gaussians.
We reproduce them using the official code to report their performance.
% We reproduce and report the Gaussian-based baselines by following the setups described in the original papers.
% We also include representatives of NeRF-based methods:
% We also report the results of several NeRF-based methods:
% K-Planes \cite{kplanes_2023}, MixVoxels \cite{wang2022mixed}, and HyperReel \cite{attal2023hyperreel}. Results of these methods are brought from  \cite{bae2024pergaussian}.
We also introduce Scaff-naive as an additional baseline.
This is an anchor-based model without our neural Gaussian design and the anchor growing, which instead adopts linear motion and the temporal opacity modeling used in \cite{yang2023gs4d}.
We present a variant of our model, Ours-light, which employs a larger voxel size to achieve a more reduced storage footprint.
% To validate the effectiveness of our proposed method, we introduce Scaff-naive as an additional baseline. Scaff-naive incorporates linear motion and a 1D Gaussian temporal opacity model into the 4D Scaffold framework.
% Furthermore, we present Ours-light, a variant of our model designed for reduced storage footprint. Ours-light achieves this by employing a larger voxel size compared to our standard approach.
% For initialization, all models utilize SfM points derived from the first frames.
For fair comparisons, we report STG trained using 300 frames and initial SfM points derived from the first frames, same as other methods.

% \textblock{Metrics.}
\myparagraph{Metrics.}
To assess the reconstruction quality, we measure peak signal-to-noise ratio (PSNR), structural similarity index (SSIM), and LPIPS \cite{zhang2018unreasonable} of the rendered images.
To compute LPIPS, we follow \cite{bae2024pergaussian} and employ AlexNet \cite{krizhevsky2012imagenet}.
To measure the visual quality in dynamic regions, we use a combined mask of Global-Median and Temporal-Difference \cite{li2022neural}, binarized with a threshold of 50.
We assess storage efficiency by calculating the total size of the output files, including MLP weights, as well as Gaussian and anchor parameters.
In addition to storage size, we also report the number of Gaussians in both dynamic and full regions.
FPS and training time are measured on a single NVIDIA A6000 GPU, using the \textit{flame\_salmon} and \textit{Fabien} scenes from N3DV and Technicolor, respectively.

\subsection{Experimental Results}
% (Our experiments obviously show dynamic scene reconstruction) \subsection{Dynamic scene reconstruction}

\myparagraph{N3DV.}
%%%% [in] 결과 다 나오고 다시 작성
\Tref{table:comparison_n3dv} presents the quantitative results on the N3DV dataset.
Our model outperforms all baselines in terms of visual quality in dynamic regions, while also delivering competitive results in full regions.
While our method employs more Gaussians to achieve high-quality results, it maintains efficient storage overhead and FPS, thanks to the anchor-based compression scheme.
Our storage-efficient variant (Ours-light) achieves the second-best storage efficiency among 4D Gaussian methods, while also surpassing other efficient baselines in visual quality.
Scaff-naive yields degraded results, indicating that the \naive extension of the scaffolding is insufficient to accurately reconstruct dynamic 3D scenes.
4DGS often models static components through multiple dynamic Gaussians, resulting in an excessive number of Gaussians and storage costs.
Although our method does not leverage additional techniques to model far-background areas \cite{li2023spacetimegaussians, yang2023gs4d}, it still exhibits plausible reconstruction quality in both static and full regions.

The effectiveness of our method is also demonstrated in qualitative results.
As shown in \Fref{fig:comparison_n3dv}, our model effectively represents dynamic regions with a sufficient number of Gaussians, thereby achieving high-quality dynamic reconstruction.
In contrast, other efficient baselines struggle to capture complex scene dynamics (see the boxed regions).
Please refer to the supplements for more visual comparisons, including uncropped results and videos.

\begin{figure}[t]
    \centering
    \includegraphics[width=1.0\linewidth, trim={0 20pt 0 0}]{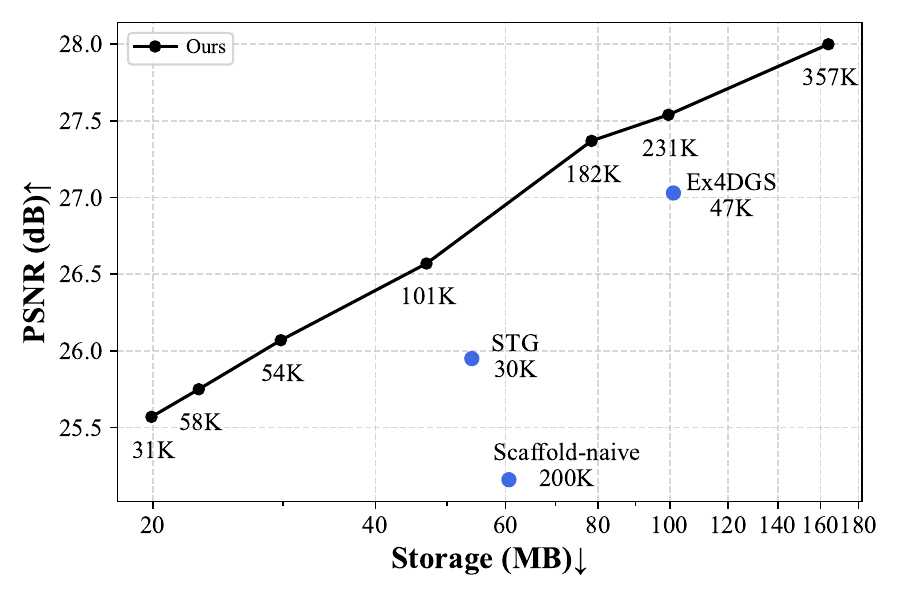}
    \caption{
    \textbf{Quality-storage trade-off}. The results on the \textit{flame\_steak} scene of N3DV, along with the number of Gaussians, are reported. We increase the number of employed Gaussians of our model by adjusting the anchor grid size.
    }
    \label{fig:analysis_storage_vs_psnr}
    % \vspace{-4mm}
\end{figure}
\begin{table}[t]
    \setlength{\tabcolsep}{4pt}
    \centering
    \footnotesize
    \begin{tabular}{ccccccc}
        \toprule
        \textbf{DA} & \textbf{Motion} & \textbf{Opacity} & \textbf{PSNR$\uparrow$} & \textbf{LPIPS$\downarrow$} & \textbf{Storage$\downarrow$} \\
        \midrule
        \checkmark & Polynomial & Ours       & 27.82 & 0.070 & 257 MB \\
        \checkmark & Linear     & 4DGS         & 27.93 & 0.065 & 195 MB \\
        \checkmark & Linear     & Ex4DGS     & 28.34 & 0.056 & 413 MB \\
                   & Linear     & Ours       & 25.77 & 0.153 & 180 MB \\
        \checkmark & Linear     & Ours       & 29.57 & 0.050 & 149 MB \\
        \bottomrule
    \end{tabular}
    \caption{\textbf{Ablation results on the main components.} ``DA'' refers to the dynamic-aware anchor growing, ``Motion''
    and ``Opacity'' for each modeling of the neural Gaussian design. }
    % \vspace{-4mm}
    \label{table:ablation}
\end{table}

\begin{figure}[t]
    \centering
    \includegraphics[width=0.95\linewidth]{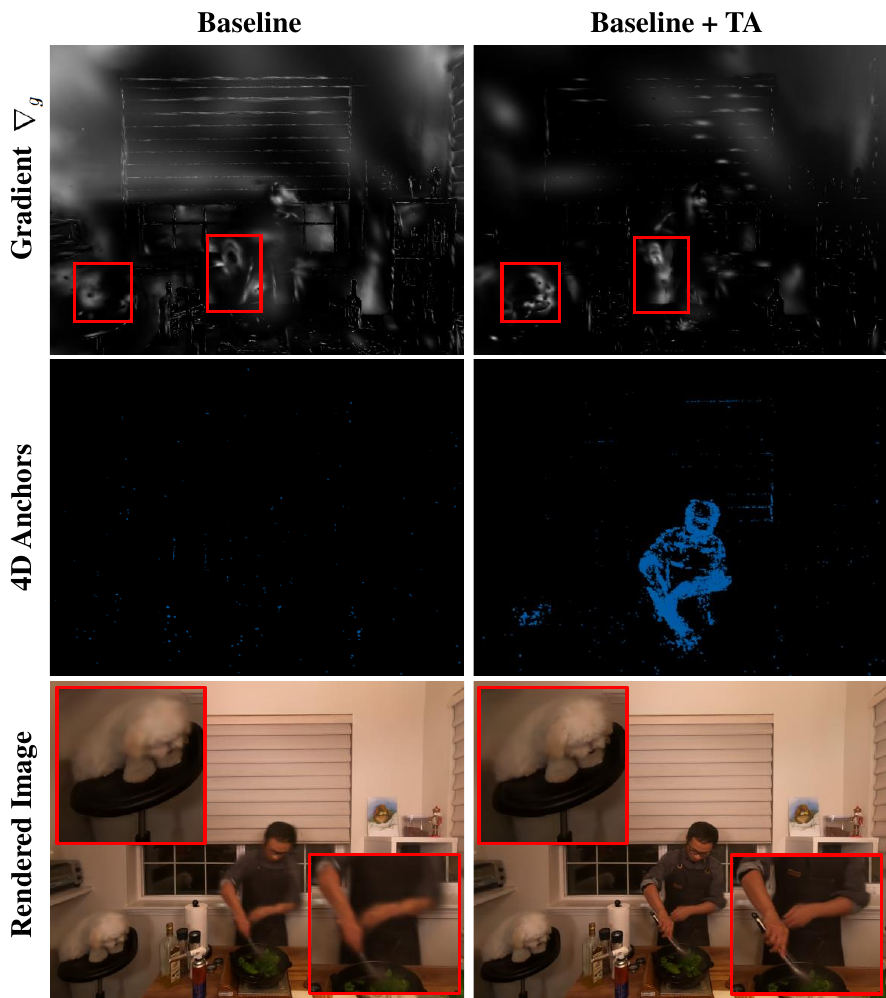}
    \caption{
    \textbf{Anchor growing analysis}. Top-to-bottom: accumulated gradients at 5000th training iteration, final 4D anchors at 106 frame, and the rendering results.
    Our anchor growing effectively accumulates gradients in under-reconstructed dynamic regions.
    % The upper row shows our vanilla model, while the lower row illustrates the model with our growing strategy applied. In the improved model, 2D Gaussian gradients are more concentrated in dynamic regions, leading to better anchor growth.
    }
    \label{fig:analysis_densification}
    % \vspace{-4mm}
\end{figure}

\myparagraph{Technicolor.}
We deliver the quantitative comparisons on the Technicolor dataset in \Tref{table:comparison_technicolor}.
Our model achieves state-of-the-art performance across all visual quality evaluation metrics.
This is further exemplified by our storage-efficient model, Ours-light.
Despite its compact size of approximately 100 MB, Ours-light delivers competitive visual quality on par with 4DGS, highlighting the effectiveness of the proposed anchor-based scheme. Qualitative comparisons on the Technicolor dataset are provided in the supplements.

% \Fref{fig:comparison_technicolor} presents the qualitative comparisons with Gaussian-based methods in the Technicolor data set.
% Our method produces sharp and high-fidelity rendering results in both static and dynamic regions.
% On the other hand, 4DGaussians, E-D3DGS and STG show blurry results in dynamic regions, such as the boxed part in the \textit{Theater} scene.
% E-D3DGS and STG often fail to reconstruct several details, \eg the eye in the \textit{Theater} scene.

\subsection{Analysis and Ablation Study}
To provide deeper insights of our main features, we conduct the ablations and analysis in dynamic regions on N3DV.
% Please refer to the supplements for more detailed analysis and ablations on each component.

\myparagraph{Quality-storage trade-off.} 
We first analyze the quality-storage trade-off of our model by gradually increasing the number of employed Gaussians.
This is controlled by adjusting the voxel size of the anchor grid.
As presented in \Fref{fig:analysis_storage_vs_psnr}, the performance of our model is improved with more Gaussians, confirming the necessity of using a sufficient quantity.
Notably, our method outperforms other storage-efficient baselines when using similar or smaller storage costs.
Thanks to the anchor-based scheme, our method employs $3.37\times$ or $7.0\times$ more Gaussians and achieves higher visual quality, yet still maintains a lower storage overhead.

\myparagraph{Understanding the anchor growing.}
We then ablate the proposed anchor growing strategy.
As reported at the bottom of \Tref{table:ablation}, the dynamic-aware anchor growing leads to noticeable improvement in visual quality.
By properly allocating new anchors to under-reconstructed dynamic regions, our model accurately represents dynamic scenes with practical storage overhead.

To further explore the effects of the anchor growing strategy, we provide visual comparisons in \Fref{fig:analysis_densification}.
The previous method fails to accurately collect gradients in dynamic regions due to their short appearing periods.
This leads to a lack of anchors in dynamic regions, resulting in degraded visual quality.
In contrast, our method successfully accumulates higher gradients in these dynamic regions (see the red-boxed parts in \Fref{fig:analysis_densification}). 
Consequently, the anchors generated by our method more accurately represent dynamic regions at each timestep. These results demonstrate that our anchor-growing strategy effectively captures under-reconstructed dynamic regions, playing a critical role in achieving high-quality reconstruction.

\myparagraph{Effects of neural Gaussian design.}
We first evaluate our neural design by replacing each modeling with the formulation in previous works.
Specifically, we replace the linear motion with the polynomial trajectory \cite{li2023spacetimegaussians}, and replace our modified temporal opacity with the one used in 4DGS \cite{yang2023gs4d} and Ex4DGS \cite{lee2024ex4dgs}.
As shown in \Tref{table:ablation}, our model with the proposed components achieves the best results.
Our compact parametrization enhances the expressiveness of each Gaussian with minimal parameters, improving both visual quality and efficiency.

\section{Conclusion}
Our framework employs a sufficient number of Gaussians to capture complex dynamic regions, while addressing the storage overhead through the anchor-based compression.
The dynamic-aware anchor growing and the neural Gaussian design lead to a substantial improvement.
Experimental results support the validity of our method, achieving state-of-the-art visual quality and practical storage costs.

% Because our method is based on \cite{lu2024scaffold}, it also shares certain limitations. We observed that the opacity at specific channel indices in the opacity MLP tends to be very low, indicating that the anchors are not fully utilized. Addressing this utilization issue could improve efficiency.

% Additionally, our method is currently optimized for multi-view video datasets, so it may not perform as effectively on monocular video scenes (e.g., D-NeRF \cite{pumarola2020d}, HyperNeRF \cite{park2021hypernerf}).
\myparagraph{Limitations and future work.}
Since our method is currently designed for multi-view video datasets, applying it to monocular videos can introduce additional challenges.
Like other methods, our method still suffers from reconstructing elements that appear very shortly (1 or 2 frames).
Resolving this challenge can be an interesting future work.
% Additionally, we observe that some anchors tend to produce only a few active Gaussians, and addressing this issue would be an interesting future avenue.

\bibliography{main}

% Check whether the conference requires a reproducibility checklist to be included in the paper.
% If so, you can uncomment the following line and ajust the path to include it.
% \input{../../ReproducibilityChecklist/LaTeX/ReproducibilityChecklist.tex}

% \clearpage
% \input{Reproducibility/ReproducibilityChecklist}

\clearpage
\setcounter{page}{1}
% \maketitlesupplementary
\maketitle
% \appendix

\renewcommand{\thetable}{S\arabic{table}}
\renewcommand{\thefigure}{S\arabic{figure}}
\setcounter{figure}{0}
\setcounter{table}{0}

% Project Page
\section*{Appendix}
% We have made our project page public and uploaded additional video results, which can be viewed at [url]. %Please take a look!
% We have made our project page public and added additional video results, which can be viewed at ProjectPage/index.html or https://fhueisbq.github.io/4dscaffold. %Please take a look!

\section{Additional Comparison with STG}
\begin{figure}[h]
    \centering
    \includegraphics[width=1\linewidth]{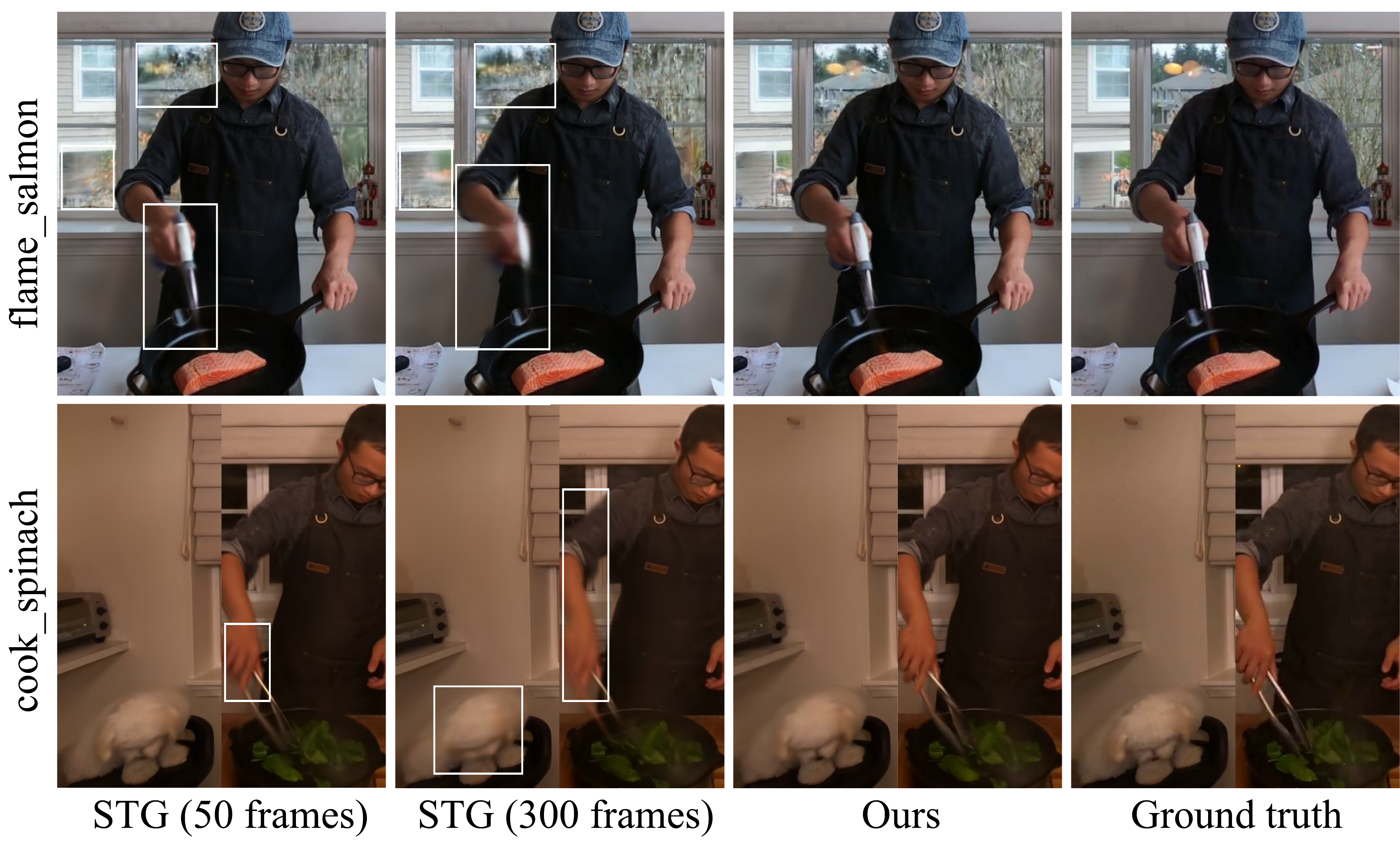}
    \caption{
    \textbf{Comparison between 50-frame STG and Ours.} The single STG model trained on 300 frames of the N3DV dataset shows lower quality in the dynamic regions compared to Ours. The white boxes highlight under-reconstruction areas.
    }
    \label{fig:supp_stg_ours}
\end{figure}

Unlike our approach, which operates from sparse initial points, STG's \cite{li2023spacetimegaussians} original setting utilizes dense SfM point clouds from all timesteps and train the scene in 50-frame sequences. This original setting requires significant preprocessing for COLMAP \cite{schoenberger2016sfm} calculation and less densification process during training. Table \ref{table:supp_stg_ours} and Figure \ref{fig:supp_stg_ours} show a qualitative and quantitative comparison between STG and our method, using the same number of iterations. While STG achieves higher scores when operating in its original dense setting, our method still demonstrates a superior score in visual quality. This indicates that our method successfully reconstructs fine-grained details that were omitted by the initial SfM points.

\begin{table}[h]
\setlength{\tabcolsep}{4pt}
\centering
\begin{tabular}{l|cc|cc}
\toprule
 & \textbf{\# models} & \textbf{\# iterations} & \textbf{PSNR}$\uparrow$ & \textbf{LPIPS}$\downarrow$ \\ \hline

\multirow{2}{*}{STG} 
 & 6 (50 frames) & 6$\times$2$^\ast$$\times$30K& 31.96 & 0.046 \\
 & 1 (300 frames) &  2$^\ast$$\times$60K & 31.25 & 0.053 \\
\midrule
Ours & 1 (300 frames) & 120K & \textbf{32.03} & \textbf{0.041} \\
\bottomrule
\end{tabular}
\caption{
\textbf{Comparison with STG on the N3DV dataset.} Our model outperforms the average of six STGs trained on 50 frames and a single STG trained on 300 frames. $\ast$ Note that STG uses a batch size of 2 while our model uses a batch size of 1.
}
\label{table:supp_stg_ours}
\end{table}

\section{Understanding Temporal Opacity}
We explore the effects of the temporal opacity by observing actual distributions.
Specifically, we choose an example pixel in \textit{sear\_steak}, which contains an object that appears in a certain period (the frying pan).
We visualize Gaussians that primarily contribute to render this pixel.
We compare our temporal opacity with the previous one based on a univariate Gaussian function.

\begin{figure}[t]
    \centering
    \includegraphics[width=0.95\linewidth]{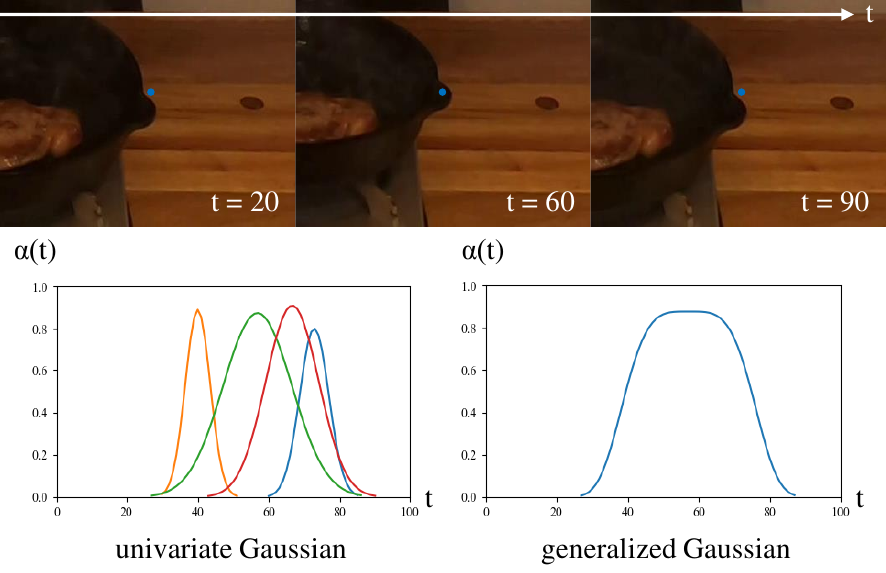}
    \caption{
    \textbf{Temporal opacity analysis.}
    % We visualize Gaussians that contributing to the rendering of a sample pixel in the \textit{sear\_steak} scene from the N3DV dataset.
    % For clarity, we filter out Gaussians distant from the frying pan based on depth and image-plane probability.
    % Our modified temporal opacity better aligns with real-world dynamics, allowing a single Gaussian to represent the element.
    We visualize Gaussians contributing to the rendering of a sample pixel in the \textit{sear\_steak} scene from the N3DV dataset.
    For clarity, we filter out Gaussians distant from the frying pan based on depth and image-plane probability
    Our modified temporal opacity better aligns with real-world dynamics, enabling a single Gaussian to represent the element.
    % The figure shows the temporal opacity distribution in a rendered scene. When an object appears and stays in place, the 1D Gaussian model needs multiple components to represent the opacity, while our generalized Gaussian achieves the same with a single component. For clarity, we filtered out distant Gaussians based on depth and image-plane probability.
    }
    % \vspace{-4mm}
    \label{fig:analysis_opacity}
\end{figure}

As shown in \Fref{fig:analysis_opacity}, the resulting Gaussian with our modified temporal opacity better fits the actual distributions of the object.
This enables our method to represent a scene element with a single Gaussian.
On the other hand, previous univariate Gaussian is not well-aligned to real-world dynamics, resulting in multiple Gaussians representing the same element.
As a result, our temporal opacity reduces the number of anchors while retaining visual quality, which leads to compact storage usage.

\section{Effects of Hyperparameters}

% \begin{figure}[tb]
\begin{figure}
    \centering
    \includegraphics[width=1\linewidth]{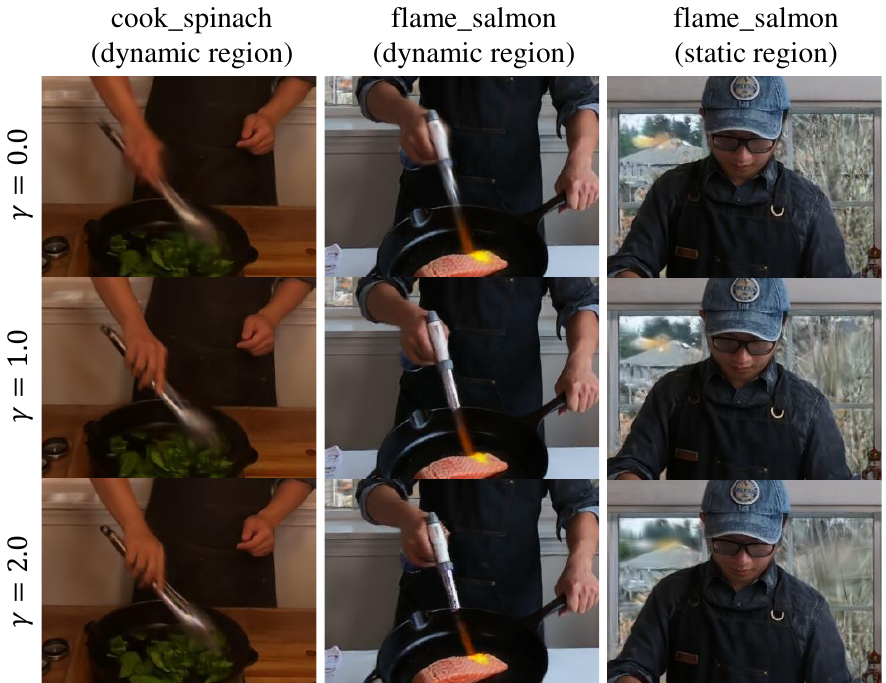}
    \includegraphics[width=1\linewidth]{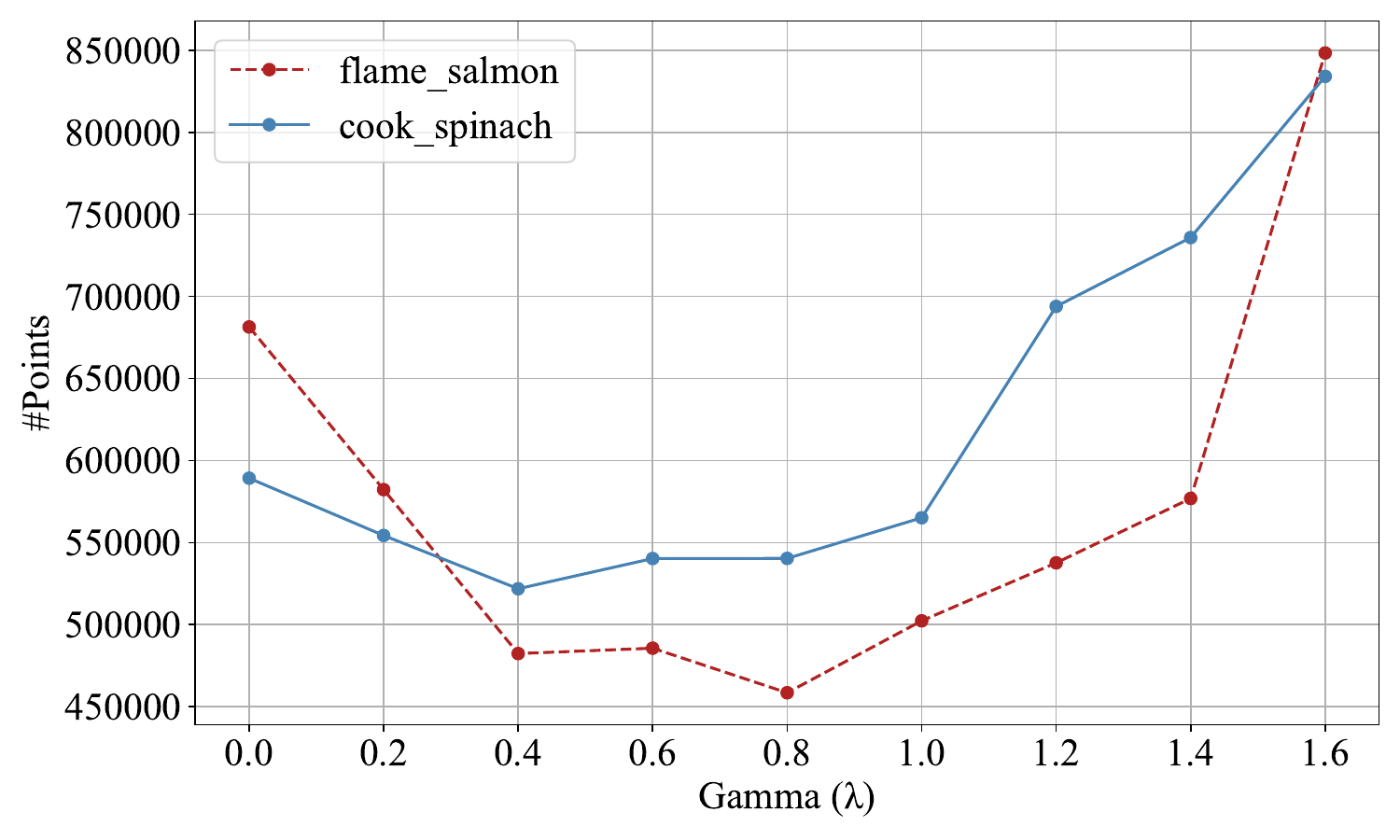}
    % \vspace{-3mm}
    \caption{
    \textbf{Effect of $\gamma$.}
    \textbf{(upper)} As the value of $\gamma$ increases, the quality of dynamic regions improves, whereas the quality of static regions declines. \textbf{(lower)} The number of anchor points changes with $\gamma$.
    }
    \label{fig:analysisgamma}
    % \vspace{20mm}
\end{figure}
% \begin{figure}[ht]
\begin{figure}
    \centering
    \includegraphics[width=1\linewidth]{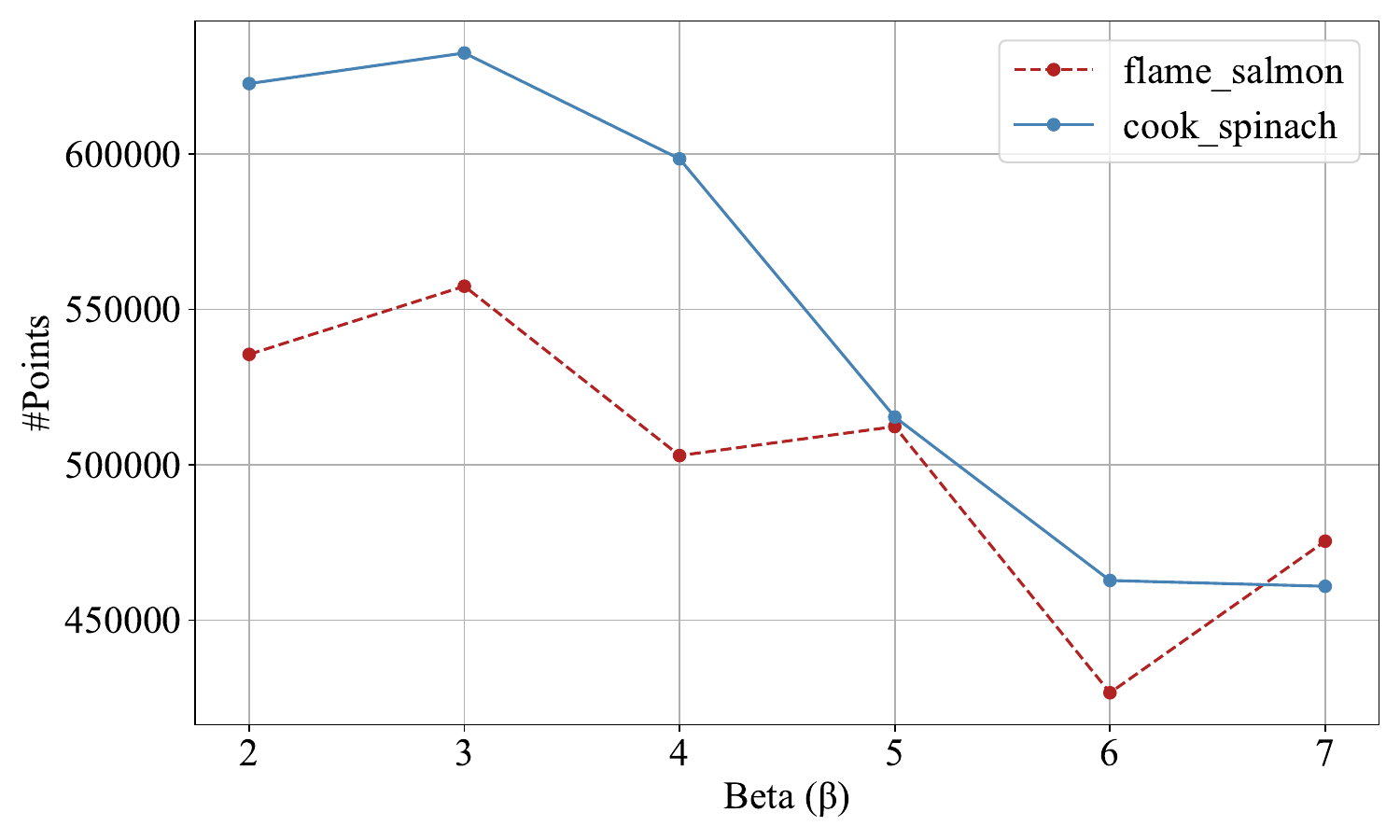}
    % \includegraphics[width=0.95\linewidth]{Supple/Assets/figure_beta_psnr_bar.pdf}
    % \vspace{-3mm}
    \caption{
    \textbf{Effect of $\beta$.}
    The graph shows how $\beta$ affects the number of anchor points across two scenes, with an increase in $\beta$ leading to a reduction in the number of anchor points.
    % The graph shows how $\beta$ influences the number of anchor points and PSNR across two scenes. Dashed lines show the trends.
    % As the value of $\beta$ increases, the number of anchor points decreases while maintaining image quality.
    }
    \label{fig:analysisbeta}
\end{figure}
\begin{figure*}%[h!]
    \centering
    \includegraphics[width=1\linewidth]{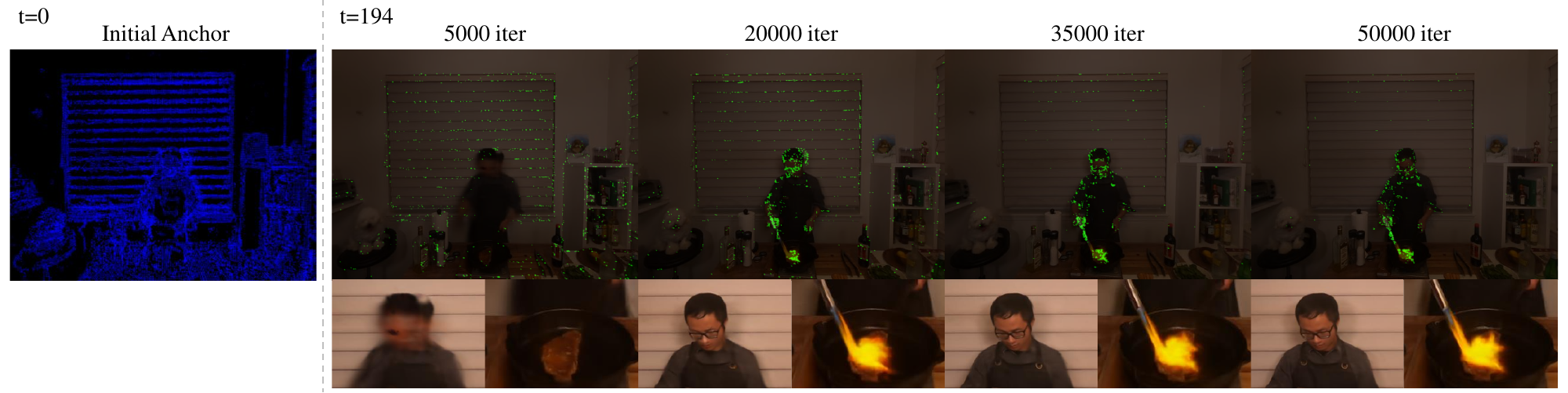}
    \caption{
    \textbf{Visualization of Anchor Growing.}  
    The blue dots on the left represent the initial anchor points, which are obtained exclusively at $t=0$. For a specific frame at $t=194$, new anchor points grow as iterations progress, shown as green dots on the right. These new anchor points are primarily generated in under-reconstructed dynamic regions.
    }
    \label{fig:anchorvis2}
    % \vspace{-5mm}
\end{figure*}

\begin{table}[h!]
\setlength{\tabcolsep}{6pt}
\centering
\begin{tabular}{c|ccccc}
\toprule
                   & $K=3$ & $K=5$ & $K=8$ & $K=10$ \\ \hline
\textbf{PSNR}      & 27.40 & 28.63 & 29.09 & 29.20  \\ 
\textbf{\#Anchors} & 579K  & 585K  & 494K  & 476K   \\
\bottomrule
\end{tabular}
\caption{\textbf{Results with various $K$.}}
\label{table:supple_noffsets}
\end{table}

We investigate the impact of hyperparameters on the performance and their controllability.
%%% [in] which scene? 명시해주면 좋을 듯
\textit{cook\_spinach} and \textit{flame\_salmon} scenes from N3DV are selected for analysis.
% : \needcheck{one representing a light scene and the other a dark scene}.

\myparagraph{Number of Gaussians per anchor.}
The parameter $K$ determines the number of Gaussians generated from an anchor feature. We observed changes in image quality and the number of anchors as $K$ varied. As $K$ increased, image quality improved while the number of anchors decreased. This is because a larger $K$ allows the same spatial region to be represented with higher density.

\myparagraph{Dynamic-aware anchor growing.}
The parameter $\gamma$ in Equation (9) controls the influence of temporal scale on densification. As $\gamma$ increases, densification becomes more concentrated in dynamic regions, improving reconstruction quality in these areas while reducing quality in static regions, as demonstrated in \Fref{fig:analysisgamma}. In scenes with distant background regions, such as \textit{flame\_salmon}, larger $\gamma$ values result in less densification in the static background, leading to a decline in reconstruction quality. Additionally, excessively high $\gamma$ values result in over-densification in dynamic regions, increasing the number of anchors.
We empirically observe that $\gamma=1$ provides a balanced reconstruction between static and dynamic regions.

\myparagraph{Temporal opacity.}
The parameter $\beta$ determines the steepness of our generalized Gaussian curve. Higher values of $\beta$ result in steeper slopes, enabling shapes to be represented with fewer Gaussian components compared to a typical mixture of univariate Gaussians. As shown in \Fref{fig:analysisbeta}, increasing $\beta$ reduces the number of required anchor points while maintaining image quality. However, excessively high $\beta$ values (e.g., above 8) cause instability during training.
 % 동적 수치로 변경하기
\section{Visualization of Anchor Growing}

We visualize the anchor growing process in \Fref{fig:anchorvis2}. During the early stages of training, the temporal scale of 4D Gaussians generated from the initial anchors increases, enabling rapid reconstruction of static regions.
As training progresses, anchors are created in the dynamic regions at each time step $t$ in under-reconstructed regions, improving image quality in these areas.
%%% [in] green dot 얘기는 caption으로 이동. 밑에 문장 위에랑 합쳐서 설명해도 될 거 같음
% \needcheck{The green dots represent the locations of newly created anchors, which are accurately placed in under-reconstructed regions.}

\section{Additional Method Details}

\myparagraph{Network architecture.}
In this section, we describe the detailed architecture of the MLPs utilized in our model. Each anchor feature is represented as a 32-dimensional vector, which is passed through the shared MLPs to compute the properties of $K$ neural 4D Gaussians. Each MLP consists of a 2-layer architecture with a hidden layer of width 32, matching the size of the feature dimension. The hidden layer uses a ReLU activation function. We employ a total of four MLPs, referred to as the Opacity MLP, Shape MLP, Color MLP, and Velocity MLP.

The Opacity MLP outputs the time-invariant opacity $\rho$ of the Gaussians. The Shape MLP produces the quaternion $\textbf{q}$ and scaling $\textbf{s}$ for covariance calculation, along with the temporal scale $\sigma$. The Color MLP takes the direction $\textbf{d}$ concatenated with the anchor feature as input and outputs the view-dependent color $\textbf{c}$. The Velocity MLP computes the 3D velocity $\textbf{u}$ of the Gaussians. $\tanh$ for opacity $\rho$, $\exp$ for scaling $\textbf{s}$ and temporal scale $\sigma$, and $\text{sigmoid}$ for color $\textbf{c}$ are applied as activation function.

\myparagraph{Additional implementation detail.}
The temporal grid size is set to 0.0333 for the N3DV dataset and 0.02 for the Technicolor dataset. For initial points, we downsample it to fewer than 100,000 points. All other hyperparameters, including the learning rate and learning decay schedule, follow those of Scaffold-GS \cite{lu2024scaffold}. Components such as the feature bank, level of detail (LOD), and appearance features are excluded.
 % Linear motion figure?
% Fewshot
% Long Video
% Longer videos. We evaluated our method on a 40 second flame salmon video. (PSNR: 28.36 dB, Storage: 634 MB) As storage grows linearly, handling even longer sequences is a promising direction for future work.

\section{Results in Challenging Scenarios}

\begin{table}[h]
\setlength{\tabcolsep}{4pt}
\centering
\begin{tabular}{c|cc|cc}
\toprule
 & \textbf{PSNR} & \textbf{SSIM} & \textbf{Train\_time} & \textbf{Storage} \\ \hline
4DGS & 24.28 & 0.833 & 4 h 30 m & 2359 MB \\ 
\textbf{Ours} & 28.82 & 0.913 & 2 h 25 m  & 114 MB  \\ 
\bottomrule
\end{tabular}
\caption{
\textbf{Results with limited views.}
}
\label{table:supple_fewshot}
\end{table}

\myparagraph{Few shots.}
To evaluate the robustness of our method, we report the results in a 4 shot scenario, in which sparse 4 views are used for reconstruction.
% We conduct the experiment with uniformly sampled 4 view from upper views in N3DV.
Results in a 4 shot scenario on the N3DV dataset are reported in \Tref{table:supple_fewshot}.
While our method was not specifically designed for few-shot scenarios, it exhibits a smaller performance drop compared to 4DGS.

% \myparagraph{Longer videos.}
% % While our experiments include a 40-second video (\textit{flame salmon}), we acknowledge that evaluations on longer videos can provide deeper insights.
% %% [웅] include 되어있다는게 어느 맥락에서 쓴 것인지? 그냥 데이터셋에 있다?
% % We will add longer-video results and discussions in the revision.
% %% [웅] long video 돌린 결과가 있긴한데, 빼기로 한 이유? 만약 안 올릴거면 이 부분은 들어내고 다른 파트를 보강하는게 좋을듯
% %%% [in] -> 아 못봤어 다시 포함하면 될듯!! -> [웅] OK
% We evaluated our method on a 40-second \textit{flame\_salmon} video. (PSNR: \textbf{28.36 dB}, Storage: \textbf{634 MB})
% % Compared to a 10-second scene, PSNR decreased by \textbf{0.84 dB} while storage increased by \textbf{4.3×}.
% % This demonstrates that our model scales to longer videos with reasonable quality and storage.
% As storage grows linearly, handling even longer sequences is a promising direction for future work.
% % Handling even longer sequences is a promising direction for future work.

\section{More Results}

\begin{figure*}[ht]
    \centering
    \includegraphics[height=0.9\textheight]{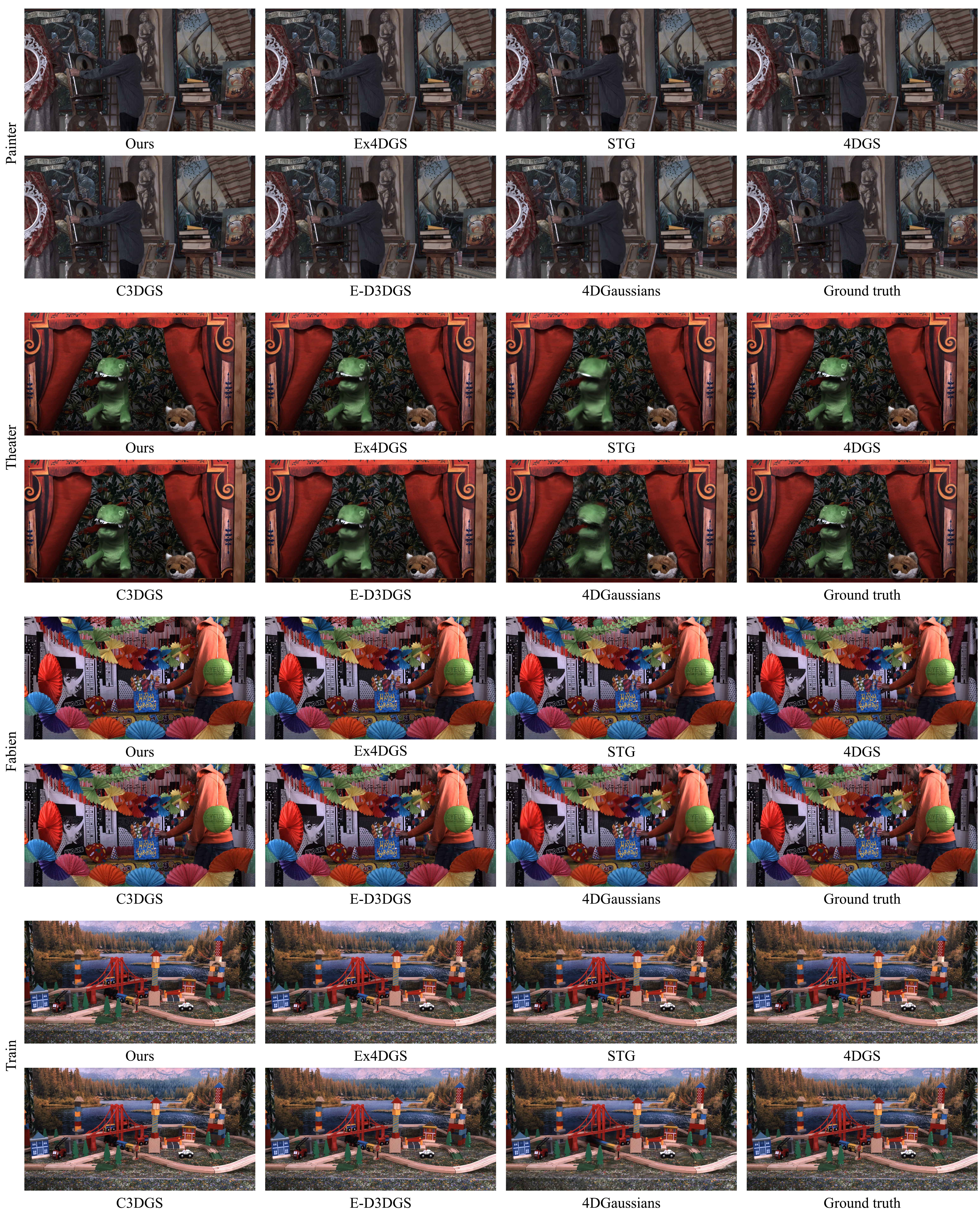}
    \caption{\textbf{Qualitative comparisons on the Technicolor dataset with full region.}}
    \label{fig:comparison_tech_supple}
\end{figure*}

We report the quantitative results in \Tref{table:supp_all_n3dv_v2}-\ref{table:supp_all_tech_v2} and qualitative results in \Fref{fig:comparison_n3dv_supple}-\ref{fig:comparison_tech_supple} of each scene of the N3DV and the Technicolor datasets, respectively. We also provide video quality comparisons in HTML format. Please check the \texttt{index.html} file.

\begin{figure*}[ht]
    \centering
    \includegraphics[width=0.9\linewidth]{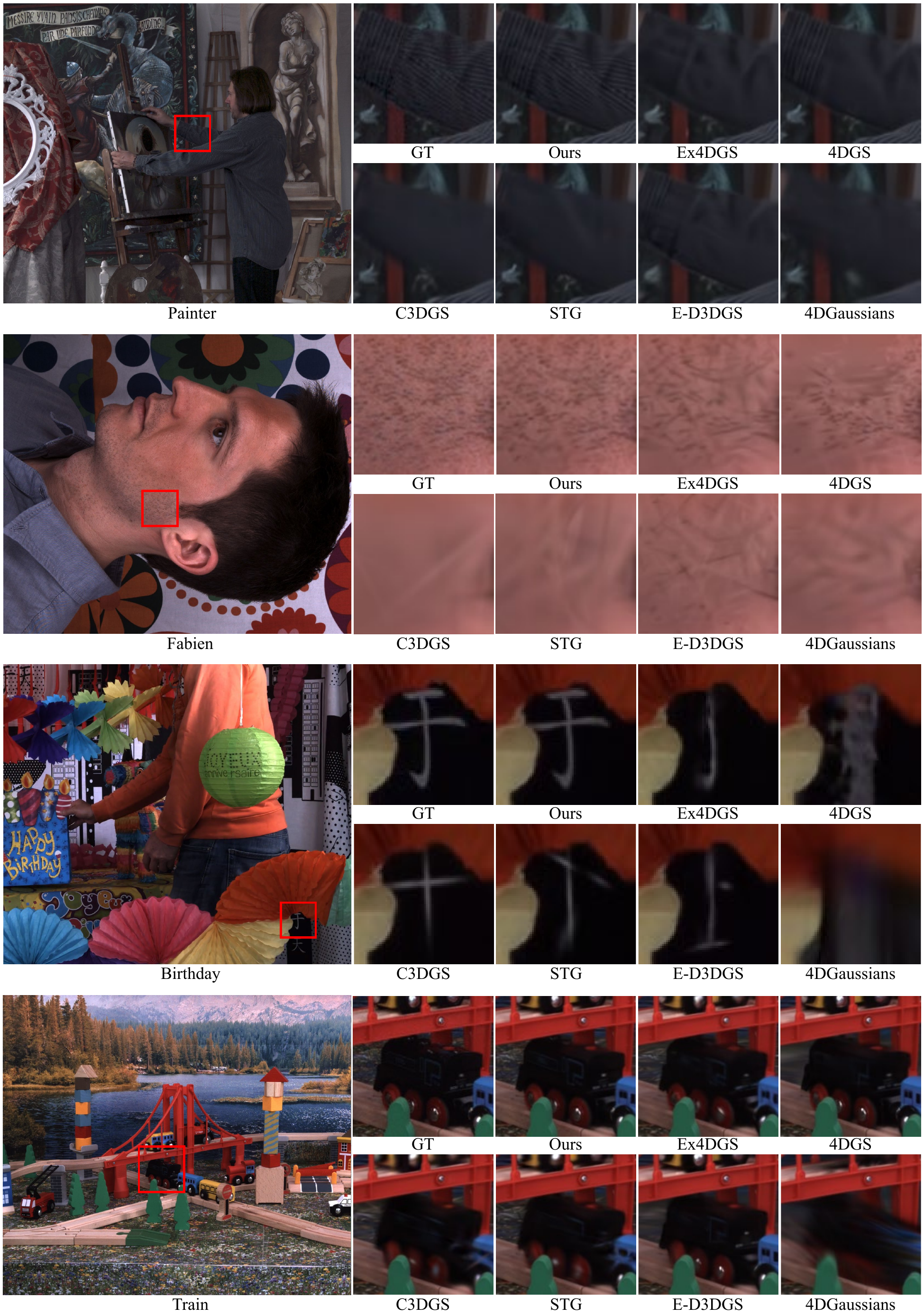}
    \caption{
    \textbf{Qualitative comparisons on the Technicolor dataset.} %This figure provides qualitative comparisons on the Technicolor dataset.
    }
    \label{fig:supple_comparison_technicolor}
    % \vspace{-2mm}
\end{figure*}

\begin{table*}[ht]
    \centering
    \resizebox{1.0\linewidth}{!}{
    \begin{tabular}{ccccccccccc}
        \toprule
        & & \multicolumn{4}{c}{Dynamic region} & \multicolumn{4}{c}{Full region} & \multicolumn{1}{c}{} \\
        \cmidrule{3-10}
        \textbf{Scene} & \textbf{Model} & \textbf{PSNR$\uparrow$} & \textbf{SSIM$\uparrow$} & \textbf{LPIPS$\downarrow$} & \textbf{\#Gaussians} & \textbf{PSNR$\uparrow$} & \textbf{SSIM$\uparrow$} & \textbf{LPIPS$\downarrow$} & \textbf{\#Gaussians} & \textbf{Storage$\downarrow$} \\
        \midrule
        \multirow{7}{*}{coffee\_martini}
        & 4DGaussian & 23.83 & 0.821 & 0.182 & - & 28.44 & 0.919 & 0.060 & 193K & 58 MB \\
        & E-D3DGS & 26.05 & 0.901 & 0.096 & - & 29.10 & 0.931 & 0.042 & 263K & 95 MB \\
        & Grid4D & 25.76 & 0.874 & 0.139 & - & 28.69 & 0.919 & 0.058 & 153K & 36 MB \\
        & STG & 25.01 & 0.884 & 0.087 & 63K & 27.45 & 0.912 & 0.070 & 481K & 70 MB \\
        & 4DGS & 26.90 & 0.920 & 0.065 & 4229K & 28.63 & 0.918 & 0.071 & 4293K & 7978 MB \\
        & Ex4DGS & 25.01 & 0.880 & 0.103 & 48K & 28.79 & 0.918 & 0.067 & 340K & 130 MB \\
        & Ours & 27.07 & 0.921 & 0.053 & 447K & 28.80 & 0.923 & 0.060 & 729K & 116 MB \\
        \midrule
        \multirow{7}{*}{cook\_spinach}
        & 4DGaussian & 24.99 & 0.783 & 0.202 & - & 33.10 & 0.953 & 0.042 & 196K & 58 MB \\
        & E-D3DGS & 26.08 & 0.824 & 0.140 & - & 32.96 & 0.956 & 0.034 & 140K & 51 MB \\
        & Grid4D & 25.88 & 0.830 & 0.163 & - & 32.90 & 0.957 & 0.035 & 231K & 55 MB \\
        & STG & 25.02 & 0.797 & 0.195 & 40K & 32.25 & 0.948 & 0.049 & 430K & 62 MB \\
        & 4DGS & 27.32 & 0.879 & 0.088 & 3514K & 33.54 & 0.956 & 0.039 & 3552K & 6599 MB \\
        & Ex4DGS & 26.10 & 0.844 & 0.137 & 60K & 33.22 & 0.955 & 0.041 & 247K & 118 MB \\
        & Ours & 27.39 & 0.878 & 0.093 & 584K & 33.36 & 0.955 & 0.036 & 782K & 198 MB \\
        \midrule
        \multirow{7}{*}{cut\_roasted\_beef}
        & 4DGaussian & 27.19 & 0.834 & 0.183 & - & 33.12 & 0.954 & 0.044 & 184K & 55 MB \\
        & E-D3DGS & 29.22 & 0.891 & 0.123 & - & 33.57 & 0.958 & 0.034 & 143K & 52 MB \\
        & Grid4D & 28.74 & 0.886 & 0.133 & - & 33.61 & 0.958 & 0.034 & 229K & 54 MB \\
        & STG & 27.87 & 0.864 & 0.150 & 38K & 32.73 & 0.951 & 0.048 & 383K & 56 MB \\
        & 4DGS & 31.21 & 0.936 & 0.060 & 3010K & 34.18 & 0.959 & 0.038 & 3047K & 5661 MB \\
        & Ex4DGS & 29.56 & 0.904 & 0.099 & 64K & 33.72 & 0.957 & 0.039 & 249K & 123 MB \\
        & Ours & 31.80 & 0.953 & 0.043 & 745K & 33.35 & 0.957 & 0.036 & 916K & 170 MB \\
        \midrule
        \multirow{7}{*}{flame\_salmon}
        & 4DGaussian & 21.98 & 0.832 & 0.163 & - & 28.80 & 0.926 & 0.060 & 197K & 58 MB \\
        & E-D3DGS & 24.29 & 0.890 & 0.104 & - & 29.61 & 0.936 & 0.038 & 264K & 96 MB \\
        & Grid4D & 24.52 & 0.888 & 0.113 & - & 29.86 & 0.930 & 0.051 & 158K & 37 MB \\
        & STG & 22.09 & 0.848 & 0.120 & 62K & 28.27 & 0.916 & 0.065 & 562K & 81 MB \\
        & 4DGS & 24.43 & 0.900 & 0.078 & 4774K & 29.25 & 0.929 & 0.057 & 4782K & 8886 MB \\
        & Ex4DGS & 22.75 & 0.860 & 0.128 & 49K & 28.78 & 0.924 & 0.064 & 330K & 128 MB \\
        & Ours & 24.82 & 0.911 & 0.059 & 473K & 29.20 & 0.928 & 0.054 & 896K & 162 MB \\
        \midrule
        \multirow{7}{*}{flame\_steak}
        & 4DGaussian & 25.43 & 0.855 & 0.134 & - & 33.55 & 0.961 & 0.032 & 186K & 56 MB \\
        & E-D3DGS & 26.54 & 0.891 & 0.101 & - & 33.57 & 0.964 & 0.028 & 135K & 50 MB \\
        & Grid4D & 25.77 & 0.880 & 0.117 & - & 32.98 & 0.963 & 0.029 & 214K & 50 MB \\
        & STG & 25.95 & 0.868 & 0.122 & 31K & 33.35 & 0.958 & 0.037 & 373K & 54 MB \\
        & 4DGS & 26.74 & 0.893 & 0.081 & 2439K & 33.90 & 0.961 & 0.037 & 2451K & 4553 MB \\
        & Ex4DGS & 27.03 & 0.905 & 0.077 & 47K & 33.90 & 0.962 & 0.032 & 226K & 101 MB \\
        & Ours & 27.84 & 0.922 & 0.049 & 500K & 33.43 & 0.962 & 0.030 & 734K & 142 MB \\
        \midrule
        \multirow{7}{*}{sear\_steak}
        & 4DGaussian & 28.59 & 0.876 & 0.127 & - & 34.02 & 0.963 & 0.031 & 198K & 59 MB \\
        & E-D3DGS & 29.36 & 0.906 & 0.107 & - & 33.45 & 0.963 & 0.030 & 135K & 49 MB \\
        & Grid4D & 29.27 & 0.905 & 0.109 & - & 33.49 & 0.965 & 0.028 & 228K & 54 MB \\
        & STG & 29.09 & 0.901 & 0.087 & 32K & 33.40 & 0.960 & 0.034 & 371K & 54 MB \\
        & 4DGS & 29.31 & 0.913 & 0.082 & 1866K & 33.37 & 0.960 & 0.040 & 1880K & 3492 MB \\
        & Ex4DGS & 30.52 & 0.927 & 0.063 & 41K & 33.68 & 0.960 & 0.034 & 221K & 94 MB \\
        & Ours & 31.44 & 0.945 & 0.045 & 448K & 33.52 & 0.960 & 0.030 & 595K & 109 MB \\
        \bottomrule
    \end{tabular}
    }
    \caption{\textbf{Quantitative results on the N3DV dataset.}}
    \label{table:supp_all_n3dv_v2}
\end{table*}
\begin{table*}[ht]
    \centering
    \resizebox{1.0\linewidth}{!}{
    \begin{tabular}{ccccccccccc}
        \toprule
        & & \multicolumn{4}{c}{Dynamic region} & \multicolumn{4}{c}{Full region} & \multicolumn{1}{c}{} \\
        \cmidrule{3-10}
        \textbf{Scene} & \textbf{Model} & \textbf{PSNR$\uparrow$} & \textbf{SSIM$\uparrow$} & \textbf{LPIPS$\downarrow$} & \textbf{\#Gaussians} & \textbf{PSNR$\uparrow$} & \textbf{SSIM$\uparrow$} & \textbf{LPIPS$\downarrow$} & \textbf{\#Gaussians} & \textbf{Storage$\downarrow$} \\
        \midrule
        \multirow{7}{*}{Birthday}
        & 4DGaussian & 25.14 & 0.838 & 0.150 & - & 28.03 & 0.862 & 0.156 & 297K & 79 MB \\
        & E-D3DGS & 29.10 & 0.948 & 0.041 & - & 33.34 & 0.951 & 0.037 & 231K & 84 MB \\
        & STG & 28.98 & 0.950 & 0.037 & 123K & 32.20 & 0.947 & 0.029 & 302K & 44 MB \\
        & C3DGS & 28.03 & 0.933 & 0.054 & 126K & 31.37 & 0.938 & 0.039 & 177K & 26 MB \\
        & 4DGS & 29.27 & 0.953 & 0.032 & 7233K & 31.72 & 0.936 & 0.042 & 7366K & 13684 MB \\
        & Ex4DGS & 29.07 & 0.949 & 0.033 & 169K & 32.25 & 0.942 & 0.031 & 460K & 162 MB \\
        & Ours & 29.51 & 0.956 & 0.026 & 395K & 32.86 & 0.952 & 0.024 & 623K & 183 MB \\
        \midrule
        \multirow{7}{*}{Fabien}
        & 4DGaussian & 28.54 & 0.807 & 0.283 & - & 33.36 & 0.865 & 0.185 & 224K & 61 MB \\
        & E-D3DGS & 30.23 & 0.859 & 0.214 & - & 34.45 & 0.875 & 0.147 & 173K & 63 MB \\
        & STG & 29.90 & 0.848 & 0.240 & 26K & 34.69 & 0.876 & 0.166 & 72K & 10 MB \\
        & C3DGS & 29.54 & 0.836 & 0.260 & 30K & 34.36 & 0.872 & 0.174 & 44K & 6 MB \\
        & 4DGS & 31.44 & 0.892 & 0.170 & 2304K & 35.02 & 0.884 & 0.144 & 2327K & 4322 MB \\
        & Ex4DGS & 31.33 & 0.890 & 0.155 & 116K & 34.98 & 0.889 & 0.131 & 202K & 83 MB \\
        & Ours & 31.44 & 0.896 & 0.151 & 380K & 35.15 & 0.895 & 0.124 & 398K & 136 MB \\
        \midrule
        \multirow{7}{*}{Painter}
        & 4DGaussian & 27.31 & 0.819 & 0.187 & - & 34.52 & 0.899 & 0.138 & 184K & 51 MB \\
        & E-D3DGS & 31.60 & 0.916 & 0.124 & - & 36.63 & 0.923 & 0.097 & 206K & 75 MB \\
        & STG & 32.10 & 0.925 & 0.120 & 61K & 36.66 & 0.923 & 0.097 & 140K & 20 MB \\
        & C3DGS & 30.39 & 0.903 & 0.145 & 61K & 35.55 & 0.913 & 0.113 & 75K & 11 MB \\
        & 4DGS & 33.51 & 0.946 & 0.097 & 4945K & 35.71 & 0.923 & 0.104 & 5015K & 9316 MB \\
        & Ex4DGS & 33.33 & 0.945 & 0.096 & 151K & 36.62 & 0.932 & 0.091 & 255K & 106 MB \\
        & Ours & 35.19 & 0.960 & 0.075 & 1729K & 37.05 & 0.939 & 0.073 & 1834K & 465 MB \\
        \midrule
        \multirow{7}{*}{Train}
        & 4DGaussian & 17.11 & 0.470 & 0.398 & - & 23.54 & 0.756 & 0.206 & 332K & 88 MB \\
        & E-D3DGS & 28.48 & 0.914 & 0.068 & - & 31.84 & 0.922 & 0.074 & 216K & 78 MB \\
        & STG & 26.56 & 0.896 & 0.110 & 101K & 32.20 & 0.935 & 0.044 & 314K & 46 MB \\
        & C3DGS & 25.23 & 0.855 & 0.174 & 138K & 30.79 & 0.910 & 0.071 & 193K & 28 MB \\
        & 4DGS & 33.09 & 0.956 & 0.034 & 6468K & 32.15 & 0.930 & 0.047 & 6980K & 12967 MB \\
        & Ex4DGS & 29.97 & 0.943 & 0.058 & 48K & 31.40 & 0.927 & 0.055 & 816K & 216 MB \\
        & Ours & 32.42 & 0.964 & 0.026 & 2339K & 32.86 & 0.946 & 0.032 & 2433K & 275 MB \\
        \midrule
        \multirow{7}{*}{Theater}
        & 4DGaussian & 21.86 & 0.612 & 0.381 & - & 28.67 & 0.835 & 0.195 & 302K & 80 MB \\
        & E-D3DGS & 27.56 & 0.794 & 0.291 & - & 30.64 & 0.864 & 0.147 & 237K & 86 MB \\
        & STG & 24.19 & 0.727 & 0.360 & 45K & 30.92 & 0.878 & 0.143 & 188K & 27 MB \\
        & C3DGS & 22.85 & 0.690 & 0.411 & 81K & 30.54 & 0.872 & 0.155 & 123K & 18 MB \\
        & 4DGS & 32.25 & 0.904 & 0.132 & 7036K & 31.90 & 0.877 & 0.136 & 7110K & 13208 MB \\
        & Ex4DGS & 29.21 & 0.862 & 0.204 & 119K & 31.74 & 0.879 & 0.131 & 399K & 132 MB \\
        & Ours & 31.12 & 0.885 & 0.188 & 980K & 32.65 & 0.893 & 0.106 & 1073K & 332 MB \\
        \bottomrule
    \end{tabular}
    }
    \caption{\textbf{Quantitative results on the Technicolor dataset.}}
    \label{table:supp_all_tech_v2}
\end{table*}

\begin{figure*}[ht]
    \centering
    \includegraphics[height=\textheight]{Supple/Assets/figure_N3DV_supple_v3.pdf}
    \caption{\textbf{Qualitative comparisons on the N3DV dataset with full region.}}
    \label{fig:comparison_n3dv_supple}
\end{figure*}

% \bibliography{main}

\end{document}